\documentclass[11pt]{article}

\usepackage[final]{acl}

\usepackage{times}
\usepackage{latexsym}

\usepackage[T1]{fontenc}

\usepackage[utf8]{inputenc}

\usepackage{microtype}

\usepackage{inconsolata}

\usepackage{graphicx}
\usepackage{amsmath}
\usepackage{amssymb}
\usepackage{booktabs}
\usepackage{tabularx}
\usepackage{array}
\usepackage{hyperref}
\usepackage{multirow}
\usepackage[dvipsnames]{xcolor}
\usepackage{wasysym}
\usepackage{tikz}
\usetikzlibrary{shapes.geometric}
\newcommand{\blackpentagon}{%
  \tikz[baseline=-0.5ex]
  \node[regular polygon, regular polygon sides=5,
        draw, fill, inner sep=0pt, minimum size=1.5ex] {};}

\title{When Is 0.1\% Enough? Analyzing the Combined Effects of Dimensionality Reduction and Quantization on Text Embedding Compression}

\author{
 \textbf{Riku Kisako}\hspace{4ex}
 \textbf{Hayato Tsukagoshi\hspace{4ex}
 \textbf{Ryohei Sasano}
}
\\
 Graduate School of Informatics, Nagoya University\\
 \{\texttt{kisako.riku.n3@s.mail, sasano@i\}.nagoya-u.ac.jp}\\
 \texttt{research.tsukagoshi.hayato@gmail.com}
}

\begin{document}
\maketitle
\begin{abstract}
Recent high-performing text embedding models often produce high-dimensional real-valued vectors, resulting in substantial storage and computational costs.
To address this issue, embedding compression methods based on dimensionality reduction or quantization have been proposed; however, the effects of their combination have not been sufficiently investigated.
In this paper, we systematically investigate how dimensionality reduction and quantization jointly affect embedding compression across four task families in the Massive Text Embedding Benchmark (MTEB), using five pretrained embedding models.
Our results show that combining dimensionality reduction and quantization can preserve near-original performance with substantially smaller storage footprints than either method alone.
In particular, for classification, several models retain at least 99\% of their original classification score using less than 0.1\% of the storage required by the original embeddings.
The preferred balance between dimensionality and bit-width varies across tasks, with retrieval generally being more sensitive to compression.
\end{abstract}

\section{Introduction}
Text embeddings are widely used as intermediate representations in NLP systems.
They map texts into vector spaces and support a range of tasks, including retrieval, classification, clustering, and semantic textual similarity (STS).
Recent text embedding models, including LLM-based models that accept task-specific instructions, have substantially improved performance across diverse benchmark tasks.
However, high-performing embedding models often produce high-dimensional real-valued vectors, which require substantial storage and increase the cost of similarity computation.
Post-hoc compression offers a practical way to reduce these costs without retraining the embedding model, provided that downstream performance is largely preserved.

\begin{figure}
    \centering
    \includegraphics[width=\columnwidth]{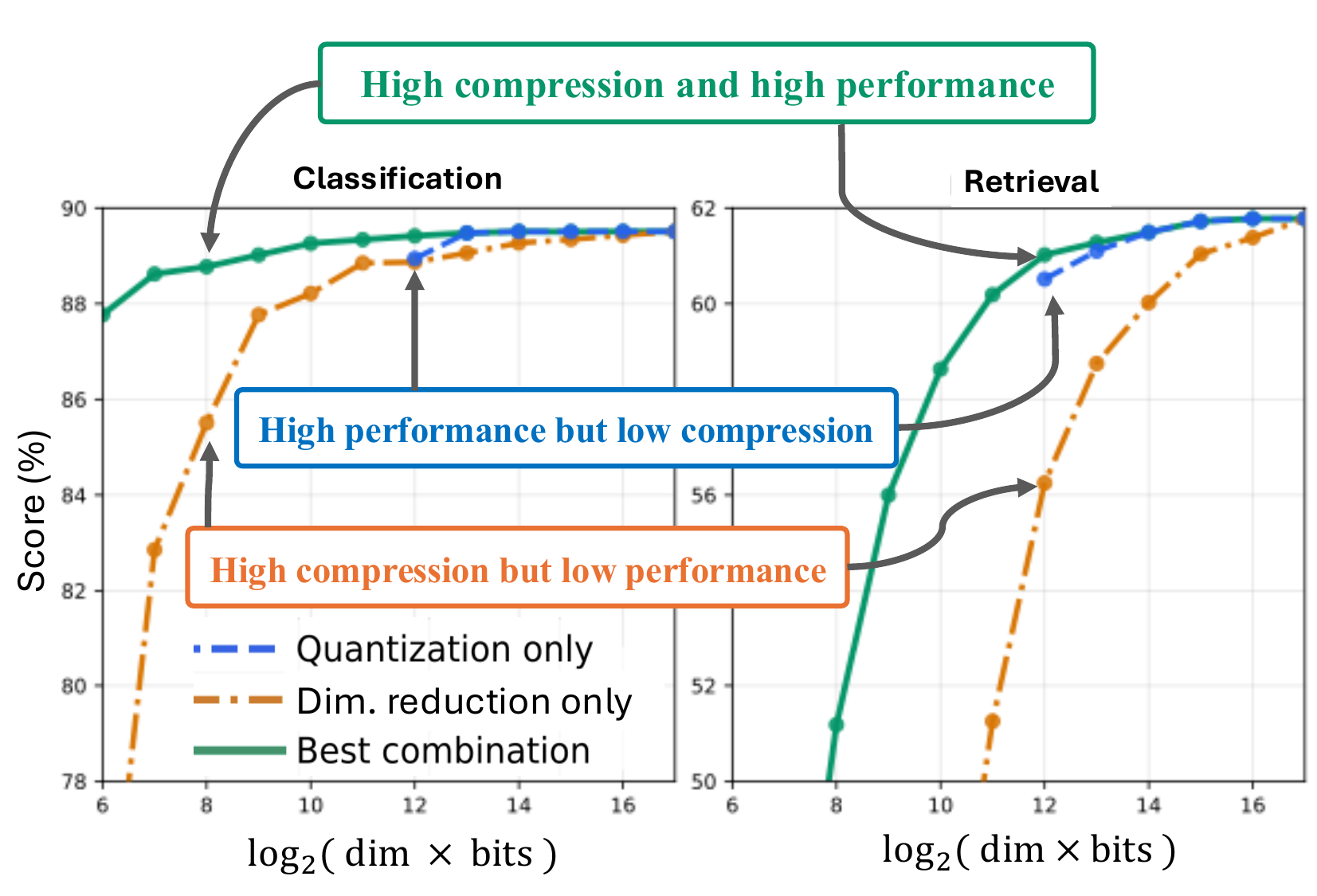}
    \vspace{-2.5ex}
    \caption{An excerpt from the experimental results for Qwen3-Embedding-8B using Head-based dimensionality reduction and quantization. The left and right panels show classification and retrieval performance, respectively, under quantization alone, dimensionality reduction alone, and their combination.}
    \vspace{-0.0ex}
    \label{fig:fig1}
\end{figure}

Two common approaches to embedding compression are quantization and dimensionality reduction. Quantization represents each coordinate with fewer bits, reducing numerical precision while keeping the dimensionality fixed. 
Dimensionality reduction instead reduces the number of retained coordinates. 
Despite progress in both approaches, their interaction remains insufficiently understood.
Because these methods control different aspects of compression, representations with the same total bit footprint per vector can differ in both dimensionality and numerical precision. A high-dimensional low-bit representation and a low-dimensional high-bit representation may therefore preserve different information and yield different downstream performance. This raises the question of how to balance dimensionality and bit-width under aggressive compression.

In this paper, we systematically evaluate the joint effects of quantization and dimensionality reduction on text embedding compression. We conduct experiments on four task families in the Massive Text Embedding Benchmark (MTEB)~\cite{muennighoff-etal-2023-mteb}: classification, clustering, retrieval, and STS. 
We use five pretrained embedding models with instruction-based or prefix-based input formats. 
Our aim is to characterize how the balance between dimensionality and bit-width affects downstream performance across tasks and models.
We show that combining even simple post-hoc methods can achieve substantial compression while preserving near-original downstream performance in some settings. 
The preferred balance between dimensionality and bit-width depends on the task and model.

Figure~\ref{fig:fig1} illustrates our key finding. 
In these results, quantization alone causes only a small performance drop, but the achievable reduction in storage is limited.
Since the original embeddings in this example use 32-bit values, converting them to 1-bit representations reduces storage to 1/32 of the original size. 
Dimensionality reduction alone, on the other hand, can achieve much stronger compression, but aggressive dimensionality reduction can lead to substantial performance degradation. 
An appropriate combination of them can preserve near-original performance at storage footprint that quantization alone cannot achieve.
For classification in Figure~\ref{fig:fig1}, when the storage per vector is reduced to 1/128 of its original size, the score decreases only from 89.51 to 89.26. 
Even at 1/1024 of the original storage, the score remains at 88.62. 
This corresponds to approximately 0.1\% of the original storage while retaining about 99\% of the original classification score.

\section{Background and Related Work}

\subsection{Text Embeddings}

Early sentence embedding methods derived sentence-level representations from word embeddings or sentence encoders~\cite{arora2017simple,conneau-etal-2017-supervised, cer2018universalsentenceencoder}. 
Later methods such as Sentence-BERT and SimCSE improved semantic representations by fine-tuning pretrained language models~\cite{reimers2019sentence,gao2021simcse}. 
Text embedding models are now evaluated across diverse task families using benchmarks such as MTEB.
Strong general-purpose embedding models have been developed using large-scale
contrastive learning, weak supervision, task-specific prefixes, or natural
language instructions~\cite{wang2022text, su-etal-2023-one,
asai-etal-2023-task, li2023towards, xiao2024c, nussbaum2024nomic,
wang2024improving, lee2025nv}. These models provide strong downstream
performance, but their high-dimensional outputs motivate efficient compression.

\subsection{Embedding Compression}

Embedding compression aims to reduce the cost of storing and comparing vectors. 
We focus on two complementary post-hoc compression approaches: quantization, which reduces the number of bits per coordinate, and dimensionality reduction, which reduces the number of coordinates.

\subsubsection{Quantization}

Quantization is widely used for neural model compression, including reduced
precision weights and activations for BERT and LLMs~\cite{zafrir2019q8bert, frantar2022gptq, xiao2023smoothquant, lin2024awq}.
It has also been applied directly to embeddings. Isotropic Iterative
Quantization compresses word embeddings into binary vectors~\cite{liao2020embedding}, while product quantization has been used to reduce
the memory and search cost of dense retrieval indexes~\cite{zhan2021jointly, qiu2022efficient}. Recent work has further examined
low-bit vector quantization for RAG and vector search~\cite{jeong20244bit, zandieh2025turboquantonlinevectorquantization}.

Hashing and rotation-based quantization have also been studied as ways to
obtain compact vector representations. \citet{charikar2002similarity}
introduced random-hyperplane locality-sensitive hashing, which maps vectors
to binary codes with collision probabilities determined by their angular
similarity. \citet{gong2011iterative} proposed Iterative Quantization (ITQ),
which learns an orthogonal rotation to minimize binary quantization error,
while \citet{ge2013optimized} optimized the space decomposition and codebooks
of product quantization to reduce distortion. For text retrieval,
\citet{yamada-etal-2021-efficient} proposed Binary Passage Retriever (BPR),
which uses binary codes for candidate retrieval and continuous vectors for
reranking. Our study complements these methods by examining the joint effects
of post-hoc dimensionality reduction and scalar quantization across text
embedding models and downstream task families under a common storage budget.

\subsubsection{Dimensionality Reduction}

Dimensionality reduction compresses embeddings by decreasing the number of
coordinates. Classical post-processing methods such as PCA, random orthogonal rotation,
Isomap, and UMAP can be applied without retraining the embedding model~\cite{abdi2010principal, achlioptas2003database, tenenbaum2000global,
mcinnes2018umap}. Another line of work trains models to support
variable-dimensional representations. Matryoshka Representation Learning (MRL)
makes prefixes of an embedding useful at multiple dimensionalities~\cite{kusupati2022matryoshka}, and this idea has been adopted in recent
embedding models such as Gemini Embedding and Qwen3-Embedding~\cite{lee2025gemini, zhang2025qwen3}.

Recent work also shows that post-hoc dimensionality reduction can preserve
performance even for models not explicitly trained with MRL.
\citet{tsukagoshi-sasano-2025-redundancy} showed that prompt-based text
embeddings are highly robust to dimensionality reduction, especially for
classification and clustering. \citet{takeshita-etal-2025-randomly} found that
randomly removing up to half of the dimensions causes only minor degradation in
retrieval and classification. Other studies have begun to evaluate
dimensionality reduction and quantization jointly in embedding storage or
compression frameworks~\cite{huergapérez2025optimizationembeddingsstoragerag,
caspari2026corect}. However, the trade-off between bit-width and dimensionality
has not been systematically characterized across task families. Our work
addresses this gap by evaluating combinations of post-hoc dimensionality
reduction and low-bit quantization under a unified compression footprint.

\section{Experimental Setup}

\subsection{Tasks and Evaluation Metrics}
We evaluate embedding compression across four task families in the MTEB: classification, clustering, retrieval, and STS. For computational efficiency, we use the same subset of English datasets as \citet{tsukagoshi-sasano-2025-redundancy}.

\paragraph{Classification}
We encode the training and evaluation texts and train a logistic regression
classifier on the labeled training embeddings. The classifier predicts labels
for the evaluation embeddings, and we measure accuracy as the fraction of
correct predictions.
We use five datasets: AmazonCounterfactualClassification for detecting
counterfactual statements in product reviews~\cite{oneill-etal-2021-wish};
AmazonPolarityClassification for binary sentiment classification of
Amazon reviews~\cite{mcauley-leskovec-2013-hidden};
AmazonReviewsClassification for predicting review ratings~\cite{keung-etal-2020-multilingual}; ImdbClassification for movie-review
sentiment classification~\cite{maas-etal-2011-learning}; and
ToxicConversationsClassification for detecting toxic comments in
Civil Comments.\footnote{\url{https://kaggle.com/competitions/jigsaw-unintended-bias-in-toxicity-classification}}

\paragraph{Clustering}
We encode the texts and group their embeddings using mini-batch k-means. We
compare the resulting clusters with the gold categories using
V-measure~\cite{rosenberg-hirschberg-2007-v}.
We use three datasets.
ArxivClusteringS2S groups paper titles by arXiv category.\footnote{\url{https://www.kaggle.com/datasets/Cornell-University/arxiv}}
RedditClustering groups post titles by subreddit, while StackExchangeClustering groups
question titles by Stack Exchange site~\cite{geigle-etal-2021-tweac}.

\paragraph{Retrieval}
We encode queries and candidate documents, rank the documents by their
similarity to each query, and evaluate whether relevant documents appear near
the top of the ranking using nDCG@10.
We use six hard-negative retrieval datasets. MIRACL retrieves relevant Wikipedia passages
for questions~\cite{zhang-etal-2022-making-miracl}, and Quora retrieves
semantically equivalent questions.\footnote{\url{https://quoradata.quora.com/First-Quora-Dataset-Release-Question-Pairs}}
HotpotQA retrieves supporting documents for multi-hop questions~\cite{yang-etal-2018-hotpotqa}; DBpedia-Entity retrieves relevant entity
descriptions~\cite{hasibi-etal-2017-dbpedia}; Natural Questions retrieves
Wikipedia passages for questions~\cite{kwiatkowski-etal-2019-natural}; and
MS MARCO retrieves passages for web search queries~\cite{nguyen-etal-2016-msmarco}.

\paragraph{STS}
We encode both sentences in each pair and compute their cosine similarity. We
then compare these similarities with human ratings using Spearman's rank
correlation.
The six yearly STS datasets, STS-12~\cite{agirre-etal-2012-semeval},
STS-13~\cite{agirre-etal-2013-sem},
STS-14~\cite{agirre-etal-2014-semeval},
STS-15~\cite{agirre-etal-2015-semeval},
STS-16~\cite{agirre-etal-2016-semeval}, and
STS-17~\cite{cer-etal-2017-semeval} measure graded semantic
similarity between sentence pairs. We also use STS-Benchmark, a
sentence-pair similarity benchmark~\cite{cer-etal-2017-semeval}, and
SICK-R, which measures compositional semantic relatedness~\cite{marelli-etal-2014-sick}.

For each dataset, we report the official MTEB main\_score. When
aggregating results, we first average scores over datasets within each task
family and then report task-family-level averages. We use the test
split when available; otherwise, we follow the official MTEB evaluation split.

\subsection{Experimental Models}
We evaluate five pretrained text embedding models. Three are prompt-based:
gte-Qwen2-7B-instruct, E5-mistral-7b-instruct~\cite{wang2024improving}, and
Qwen3-Embedding-8B~\cite{zhang2025qwen3}. The other two are prefix-based:
E5-large-v2~\cite{wang2022text} and
nomic-embed-text-v1.5~\cite{nussbaum2024nomic}.
Qwen3-Embedding-8B and nomic-embed-text-v1.5 support variable-dimensional
embeddings through Matryoshka Representation Learning (MRL). This allows us to
examine how models designed for variable-dimensional embeddings behave under
dimensionality reduction and quantization. More details about
the evaluated models and their input formats are provided in
Appendix~\ref{app:models}.
The dataset-specific instructions used for each task are provided in
Appendix~\ref{app:instructions}. For each model, we only evaluate target
dimensions that do not exceed its original embedding dimensionality.

\subsection{Dimensionality Reduction Methods}

Let \(f(x) \in \mathbb{R}^{D}\) be the original embedding of input \(x\).
We compare two dimensionality reduction methods that produce
\(z(x) \in \mathbb{R}^{d}\).

\paragraph{{Head}}
This baseline keeps the first \(d\) coordinates:
\[
    z(x) = f(x)_{1:d}.
\]
It requires no fitted projection.

\paragraph{{PCA+ROR}}
We use PCA-based dimensionality reduction followed by a deterministic random
orthogonal rotation. Let \(\mu \in \mathbb{R}^{D}\) be the mean vector of the
calibration embeddings, and let \(U_d \in \mathbb{R}^{D \times d}\) be the
matrix whose columns are the top \(d\) principal components. A standard PCA
projection first maps the embedding to
\[
    \tilde{z}(x) = (f(x) - \mu) U_d .
\]
However, the coordinates of \(\tilde{z}(x)\) are ordered by explained variance,
which can concentrate most of the variance in a small number of dimensions.
This coordinate-wise variance imbalance can be undesirable when the reduced
embeddings are subsequently quantized, because scalar quantization is applied
independently to each coordinate.

To mitigate this issue, we multiply the PCA-projected vector by a deterministic
random orthogonal matrix:
\[
    z(x) = (f(x) - \mu) U_d R_d,
\]
where \(R_d \in \mathbb{R}^{d \times d}\) is obtained by QR decomposition of a
Gaussian random matrix. The main experiments use projection seed 0, with
generator seed \(0+d\) for each target dimension. Since \(R_d\) is orthogonal, this
rotation preserves Euclidean distances and inner products within the PCA
subspace, while redistributing the variance across coordinates. Throughout the
main experiments, we refer to this method as PCA+ROR (PCA followed by random orthogonal rotation). Results for
standard PCA without the random orthogonal rotation are discussed separately in
Section~\ref{sec:why_random_projection}.
For each model--dataset pair, we fit PCA without task labels using 20\% of
the cached embeddings. If this sample is insufficient to fit PCA, we use
more than 20\%.

\subsection{Quantization}
After dimensionality reduction, we apply post-training embedding quantization.
All compressed embeddings are dequantized back to float32 before
evaluation, allowing us to simulate compressed storage while keeping the MTEB
evaluation pipeline unchanged.

We evaluate six bit-width settings, with \(b\) bits per scalar:
\[
    b \in \{1, 2, 4, 8, 16, 32\}.
\]
The 32-bit setting uses the original float32 values. The 16-bit
setting casts embeddings to float16 and then converts them back to
float32. The 1-bit setting thresholds each scalar at zero and maps it
to 0 or 1.

For 8-, 4-, and 2-bit quantization, we do not use low-precision floating-point
formats directly. The embedding models used in our experiments output
L2-normalized embeddings. Consequently, especially in high-dimensional
embeddings, individual scalar values tend to have very small magnitudes and are
highly concentrated around zero. Under coarse low-bit floating-point
representations, many such values can collapse to zero or to a small number of
representable levels, leaving much of the available bit budget ineffective for
distinguishing embedding values. To avoid this, we use a distribution-adaptive
lookup-table quantizer that allocates the available \(2^b\) levels according to
the empirical distribution of embedding values.

Specifically, for each model, task, dimensionality-reduction method, target
dimension \(d\), and bit-width \(b\), we construct a global equal-count scalar
quantizer. We first flatten the transformed calibration embeddings into a
single set of scalar values, sort them, and divide them into \(2^b\) bins with
approximately equal numbers of values. Each bin is
represented by the mean of the values assigned to that bin. At quantization
time, each scalar is replaced by the representative value of the bin to which it
belongs. For each model--dataset pair, we fit the quantization tables using
20\% of the cached embeddings, or more if needed to fit PCA. We evaluate on the full selected MTEB split,
so some evaluation inputs are also used to fit PCA and construct the
quantization tables. Appendix~\ref{app:evaluation_overlap} tests whether
using these inputs for fitting affects the results. Appendix~\ref{app:quant_format} compares our quantizer with fixed
quantization formats.

\subsection{Storage Footprint}
\label{sec:compression_budget}

We measure the nominal per-vector storage cost of each compressed embedding
by the number of stored bits $d\times b$, where $d$ is the reduced dimensionality and $b$ is the bit-width.

This footprint counts only the bits assigned to each stored embedding vector.
For the equal-count quantizer, the lookup table and bin boundaries introduce
additional global storage. However, these parameters are shared by all
embeddings in the same task, dimensionality, and bit-width setting. If a
quantization table with \(2^b\) representative values and its bin boundaries
are stored in 32-bit floating point, the additional cost is \(O(2^b)\) bits
per configuration, and its amortized cost per embedding decreases as
\(O(2^b/N)\), where \(N\) is the number of stored embeddings. Therefore, for
large embedding collections, this overhead is negligible compared with the
per-vector cost \(db\).

\begin{figure*}[t]
    \centering

    \begin{minipage}[t]{0.48\textwidth}
        \centering
        \includegraphics[width=\linewidth]{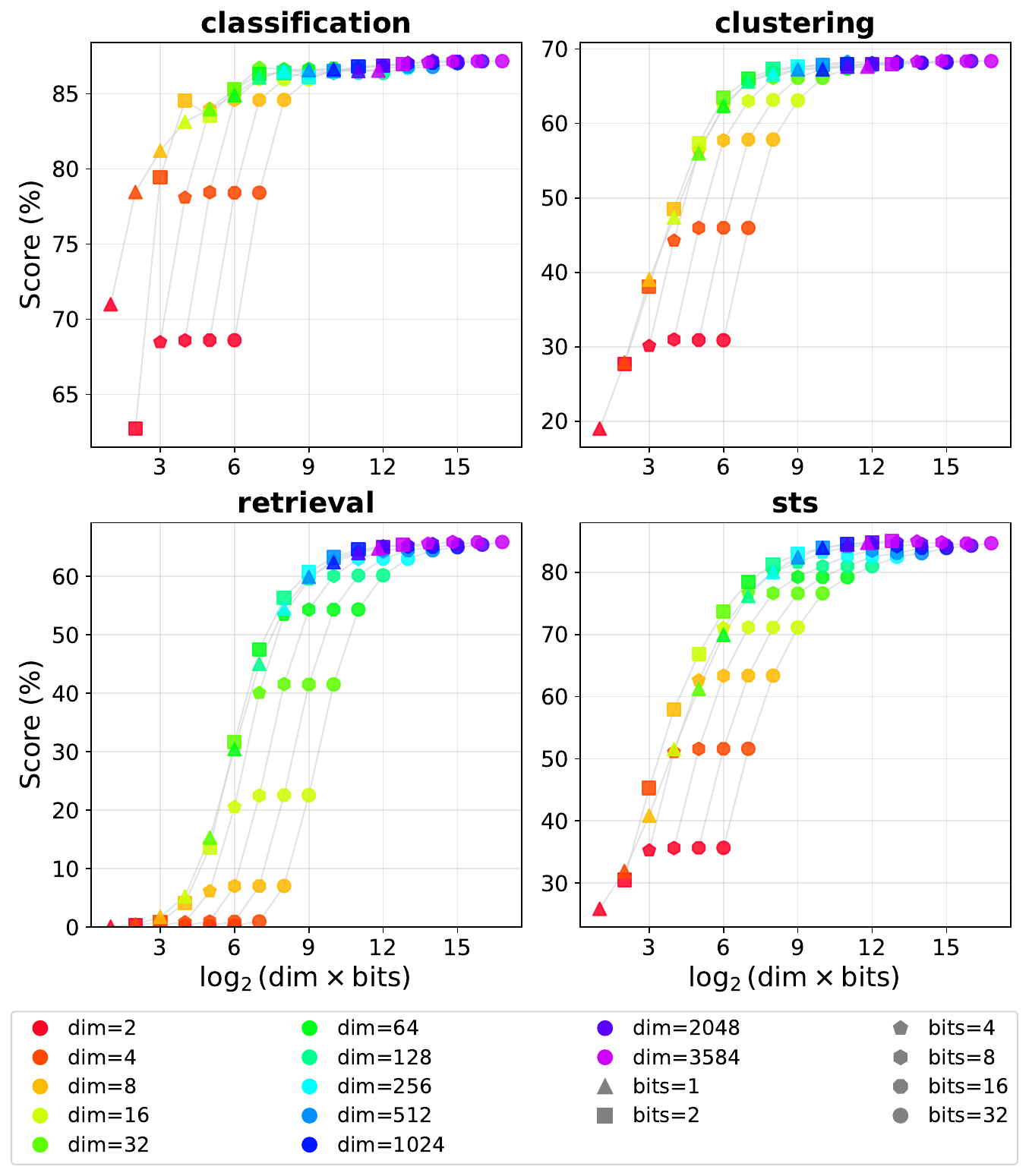}
        \caption{Performance of Head-based dimensionality reduction for gte-Qwen2 across different bit-widths and task types.
        The horizontal axis is logarithmic. The dot shapes represent the number of bits, while the dot colors indicate the number of dimensions. Results with the same number of bits are connected by gray lines.}
        \label{fig:head_gte-Qwen2}
    \end{minipage}
    \hfill
    \begin{minipage}[t]{0.48\textwidth}
        \centering
        \includegraphics[width=\linewidth]{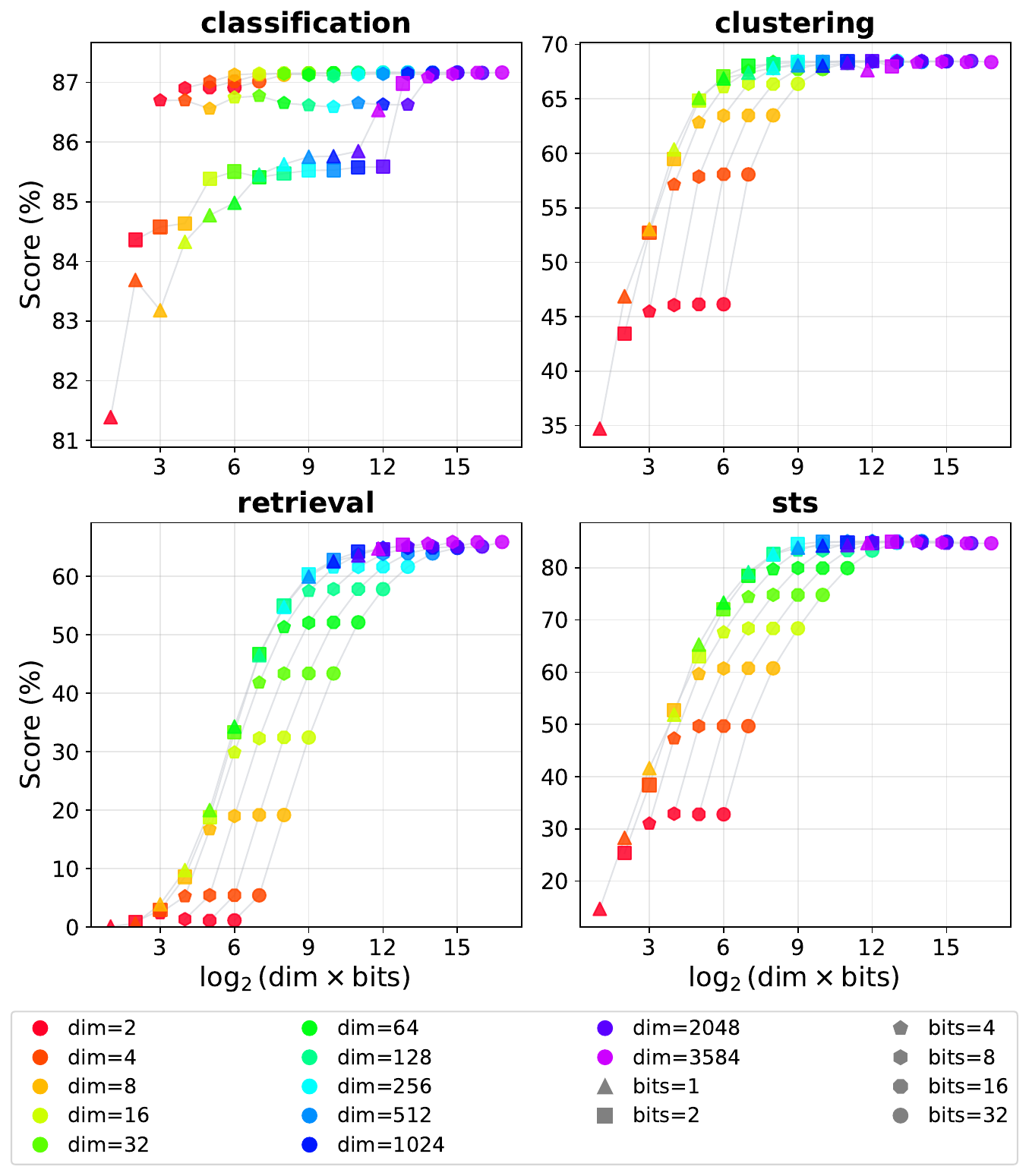}
        \caption{Performance of PCA+ROR for gte-Qwen2 across different bit-widths and task types.
        The horizontal axis is logarithmic. The dot shapes represent the number of bits, while the dot colors indicate the number of dimensions. Results with the same number of bits are connected by gray lines.}
        \label{fig:pca_random_gte-Qwen2}
    \end{minipage}
        \vspace{3.5ex}
\end{figure*}

\section{Experiments}
\label{sec:results}
We analyze how compressing embedding representations through dimensionality reduction and quantization affects downstream task performance.

\subsection{Overview of Results}
Figures~\ref{fig:head_gte-Qwen2} and ~\ref{fig:pca_random_gte-Qwen2} show the results for gte-Qwen2 using Head and PCA+ROR, respectively.
We find that combining dimensionality reduction and quantization enables greater compression while preserving performance than applying either method alone.

For example, consider the top-left panel of Figure 2, which shows classification performance using gte-Qwen2 embeddings compressed by Head-based dimensionality reduction. When the dimensionality is fixed at its original value of 3,584 (shown in {\color[rgb]{0.8,0,1}magenta}), quantization alone can reduce the storage footprint only to approximately \(\log_2(d \times b) = 12\). 
When the bit width is fixed at 32 (indicated by {\small \color{gray} $\CIRCLE$}), reducing the dimensionality to 8 (shown in {\color[rgb]{1,0.75,0.125}orange}) or fewer (corresponding to a footprint of \(\log_2(d \times b) = 8\)) leads to a noticeable performance degradation. 
In contrast, combining dimensionality reduction with quantization allows the performance to be largely preserved even with a bit width of 4 (indicated by {\small \color{gray} $\blackpentagon$}) and a dimensionality of 32 (shown in {\color[rgb]{0.53,1,0.25}lime green}), corresponding to a substantially smaller footprint of \(\log_2(d \times b) = 7\).

Table~\ref{tab:main_budget_performance} summarizes the best observed scores for gte-Qwen2, Qwen3-Embedding, and Nomic within four storage limits: 10\%, 1\%, 0.1\%, and 0.01\% of the original embedding size.
The results show that performance retained at the same relative storage budget depends on the model as well as the task.
Results for all five models are provided in Appendix~\ref{app:budget_performance} (Table~\ref{tab:budget_performance}); the other models' figures are in Appendix~\ref{app:all_results}.

\begin{table*}[t!]
\centering
\small
\setlength{\tabcolsep}{3pt}
\begin{tabular}{@{}lll r@{\hspace{0.4em}}l r@{\hspace{0.4em}}l r@{\hspace{0.4em}}l r@{\hspace{0.4em}}l@{}}
\toprule
Model & Budget & Method & \multicolumn{2}{c}{Classification} & \multicolumn{2}{c}{Clustering} & \multicolumn{2}{c}{Retrieval} & \multicolumn{2}{c}{STS} \\
\midrule
\multirow{8}{*}{gte-Qwen2} & \multirow{2}{*}{0.01\%} & Head & 93.2 & ($8\times1$) & 57.1 & ($8\times1$) & 2.6 & ($8\times1$) & 53.5 & ($4\times2$) \\
 &  & PCA+ROR & 99.5 & ($2\times4$) & 77.6 & ($8\times1$) & 5.9 & ($8\times1$) & 49.1 & ($8\times1$) \\
\cmidrule{2-11}
 & \multirow{2}{*}{0.1\%} & Head & 97.8 & ($32\times2$) & 92.9 & ($32\times2$) & 48.0 & ($32\times2$) & 87.1 & ($32\times2$) \\
 &  & PCA+ROR & 100.0 & ($8\times8$) & 98.1 & ($32\times2$) & 52.1 & ($64\times1$) & 86.6 & ($64\times1$) \\
\cmidrule{2-11}
 & \multirow{2}{*}{1\%} & Head & 99.5 & ($32\times4$) & 99.3 & ($256\times4$) & 96.0 & ($512\times2$) & 99.1 & ($1024\times1$) \\
 &  & PCA+ROR & 100.0 & ($64\times16$) & 100.1 & ($128\times4$) & 95.3 & ($512\times2$) & 100.4 & ($512\times2$) \\
\cmidrule{2-11}
 & \multirow{2}{*}{10\%} & Head & 99.9 & ($2048\times4$) & 99.9 & ($512\times4$) & 99.2 & ($3584\times2$) & 100.4 & ($3584\times2$) \\
 &  & PCA+ROR & 100.0 & ($256\times16$) & 100.2 & ($256\times32$) & 98.9 & ($2048\times4$) & 100.5 & ($512\times8$) \\
\midrule
\multirow{8}{*}{Qwen3-Emb.} & \multirow{2}{*}{0.01\%} & Head & 92.1 & ($4\times2$) & 37.6 & ($4\times2$) & 2.0 & ($8\times1$) & 54.3 & ($4\times2$) \\
 &  & PCA+ROR & 99.4 & ($2\times4$) & 66.7 & ($8\times1$) & 6.6 & ($4\times2$) & 51.8 & ($8\times1$) \\
\cmidrule{2-11}
 & \multirow{2}{*}{0.1\%} & Head & 99.0 & ($64\times2$) & 92.9 & ($64\times2$) & 70.7 & ($64\times2$) & 94.6 & ($32\times4$) \\
 &  & PCA+ROR & 100.1 & ($8\times8$) & 99.3 & ($64\times2$) & 71.6 & ($128\times1$) & 93.2 & ($128\times1$) \\
\cmidrule{2-11}
 & \multirow{2}{*}{1\%} & Head & 99.7 & ($512\times2$) & 99.0 & ($512\times2$) & 95.1 & ($512\times2$) & 99.3 & ($256\times4$) \\
 &  & PCA+ROR & 100.1 & ($8\times8$) & 100.7 & ($128\times8$) & 95.9 & ($512\times2$) & 98.6 & ($512\times2$) \\
\cmidrule{2-11}
 & \multirow{2}{*}{10\%} & Head & 100.0 & ($4096\times2$) & 100.4 & ($2048\times4$) & 99.1 & ($3584\times2$) & 100.0 & ($1024\times8$) \\
 &  & PCA+ROR & 100.1 & ($8\times8$) & 100.7 & ($128\times8$) & 98.9 & ($4096\times2$) & 98.9 & ($1024\times4$) \\
\midrule
\multirow{8}{*}{Nomic} & \multirow{2}{*}{0.01\%} & Head & 63.5 & ($2\times1$) & 6.9 & ($2\times1$) & 0.4 & ($2\times1$) & 23.6 & ($2\times1$) \\
 &  & PCA+ROR & 88.7 & ($2\times1$) & 46.6 & ($2\times1$) & 0.2 & ($2\times1$) & 18.0 & ($2\times1$) \\
\cmidrule{2-11}
 & \multirow{2}{*}{0.1\%} & Head & 69.7 & ($8\times2$) & 28.3 & ($16\times1$) & 5.5 & ($16\times1$) & 62.3 & ($4\times4$) \\
 &  & PCA+ROR & 96.5 & ($4\times4$) & 69.6 & ($8\times2$) & 12.0 & ($16\times1$) & 62.7 & ($16\times1$) \\
\cmidrule{2-11}
 & \multirow{2}{*}{1\%} & Head & 86.4 & ($64\times2$) & 82.2 & ($64\times2$) & 66.0 & ($64\times2$) & 93.9 & ($64\times2$) \\
 &  & PCA+ROR & 100.2 & ($16\times8$) & 95.2 & ($64\times2$) & 63.4 & ($128\times1$) & 93.8 & ($64\times2$) \\
\cmidrule{2-11}
 & \multirow{2}{*}{10\%} & Head & 99.5 & ($512\times4$) & 99.2 & ($512\times4$) & 98.9 & ($512\times4$) & 99.8 & ($256\times8$) \\
 &  & PCA+ROR & 100.2 & ($16\times8$) & 100.1 & ($256\times8$) & 97.7 & ($512\times4$) & 100.4 & ($512\times2$) \\
\bottomrule
\end{tabular}
\caption{Best observed gte-Qwen2, Qwen3-Embedding, and Nomic performance within fixed embedding storage budgets. Scores are percentages of each model's original full-dimensional 32-bit Head score. Parentheses show the selected dimensionality and bit-width $(d\times b)$; the budget is $100db/(32D)$, where $D$ is the model's original dimensionality. Values above 100\% exceed the reference score.}

\label{tab:main_budget_performance}
\end{table*}

\subsection{Differences in Trends by Task}
\label{sec:task_differences}
The effects of dimensionality reduction and quantization tend to vary depending on the task being processed.
\paragraph{Classification}
Classification performance remains close to the original even under
substantial compression. Table~\ref{tab:main_budget_performance} shows that
gte-Qwen2 reaches 99.5\% of its original score with PCA+ROR within a 0.01\%
storage budget, while Nomic reaches 96.5\% at a 0.1\% budget.
Qwen3-Embedding, E5-mistral, and E5-large also reach at least 99\% with
configurations using less than 0.1\% of their original storage. Thus, classification
is relatively robust to compression across the evaluated models. The
Classification panel of
Figure~\ref{fig:pca_random_gte-Qwen2} also shows that increasing the number
of dimensions does not necessarily improve performance at a fixed storage
budget. For gte-Qwen2 with PCA+ROR, $2\times4$ outperforms $8\times1$ in
classification, although both use eight bits per vector; the reverse holds
for this pair of configurations in clustering, retrieval, and STS.
\paragraph{Clustering}
For clustering tasks, the three high-dimensional models retain 99\% of their
original performance with a relative compression footprint below 1\% using at
least one reduction method. In Table~\ref{tab:main_budget_performance},
gte-Qwen2 reaches 98.1\% at a 0.1\% budget with PCA+ROR, whereas Nomic
reaches 69.6\% and needs a 10\% budget to exceed 99\%. E5-large also
requires a larger fraction of its original footprint. The relationship between
bits and dimensions is dominated by dimensionality. When comparing performance at the same
compression footprint, configurations with larger dimensionality generally show
higher scores.

\paragraph{Retrieval}
Among the four task families, retrieval generally exhibited the largest relative performance degradation under embedding compression.
For gte-Qwen2, classification, clustering, and STS maintain
nearly flat performance down to 256 dimensions, whereas retrieval already
shows degradation at 1024 dimensions. In
Table~\ref{tab:main_budget_performance}, its best retrieval score is only
5.9\% at a 0.01\% storage budget and 52.1\% at 0.1\%; it recovers to 96.0\%
at 1\%. Nomic retains just 0.4\% and 12.0\% at the two smallest storage limits,
and recovers to 98.9\% at 10\%. Across all five models, none retains even
half of its original retrieval score at a 0.01\% budget.
\paragraph{STS}
For STS tasks, the three high-dimensional models retain 99\% of their
original performance with a relative compression footprint below 1\% using at
least one reduction method, while E5-large and Nomic require larger fractions.
The relationship between bits and dimensions is relatively sensitive to
dimensionality. Like clustering and retrieval, when comparing performance at
the same compression footprint, configurations with larger dimensionality generally
achieve higher scores.

\subsection{Effect of Dimensionality Reduction Methods}
\label{sec:dim_red_methods}

Table~\ref{tab:main_budget_performance} compares the best observed scores for
Head and PCA+ROR under the same storage budgets. Each entry selects the best
evaluated dimensionality and bit-width for its method. At 0.01\%, PCA+ROR outperforms Head in classification and clustering for all three models, whereas Head performs better in STS for all three. In retrieval, PCA+ROR leads for gte-Qwen2 and Qwen3-Embedding, while Head leads for Nomic. The preferred method therefore depends on the model, task, and budget.

Under severe compression, these differences can be substantial. For
gte-Qwen2 at 0.01\%, PCA+ROR retains 77.6\% of the original clustering score,
compared with 57.1\% for Head. For Nomic at 0.1\%, the corresponding scores
are 69.6\% and 28.3\%. Thus, PCA+ROR can yield a substantially higher score
even when both methods have the same storage limit.

At larger storage limits, the advantage of PCA+ROR becomes less uniform. At 1\%,
Head slightly exceeds PCA+ROR for gte-Qwen2 retrieval (96.0\% versus
95.3\%). At 10\%, their retrieval scores are 99.2\% and 98.9\%, respectively.
Thus, the method with the highest score under aggressive compression does not
necessarily remain preferable as the budget increases.

Figures~\ref{fig:head_gte-Qwen2} and~\ref{fig:pca_random_gte-Qwen2}
illustrate these trends for gte-Qwen2. PCA+ROR achieves higher scores in
classification and clustering for configurations with small storage footprints, whereas Head is slightly
stronger in STS. For retrieval, both methods improve sharply as more
storage is allocated, and their scores are close at 1\%. Overall, the
relative performance of the two methods varies with the task and the
available storage budget.

\subsection{Model-Dependent Compression Robustness}
\label{sec:original_dimensionality}
At 128 bits under Head, gte-Qwen2 (3584 original dimensions) retains 99.5\%
of its classification score with $32\times4$, versus 86.4\% for Nomic (768
dimensions) with $64\times2$ (Table~\ref{tab:main_budget_performance}). This
suggests original dimensionality may matter, although the selected
dimension--bit-width balances and model training also differ.

\label{sec:matryoshka_compatibility}
At a 1\% storage budget under Head, the MRL-compatible Qwen3-Embedding
retains 99.0\% in clustering and 95.1\% in retrieval, compared with 82.2\%
and 66.0\% for MRL-compatible Nomic. Non-MRL gte-Qwen2 retains 99.3\% and
96.0\% for the same tasks. Thus, MRL compatibility alone does not explain
robustness. This is consistent with \citet{takeshita2026mrl}, who found that
non-MRL text encoders can match or outperform MRL encoders under moderate
prefix truncation, while MRL can help under heavier truncation.

\subsection{Effect of Random Orthogonal Rotation after PCA}
\label{sec:why_random_projection}
\begin{figure}
    \centering
    \includegraphics[width=\linewidth]{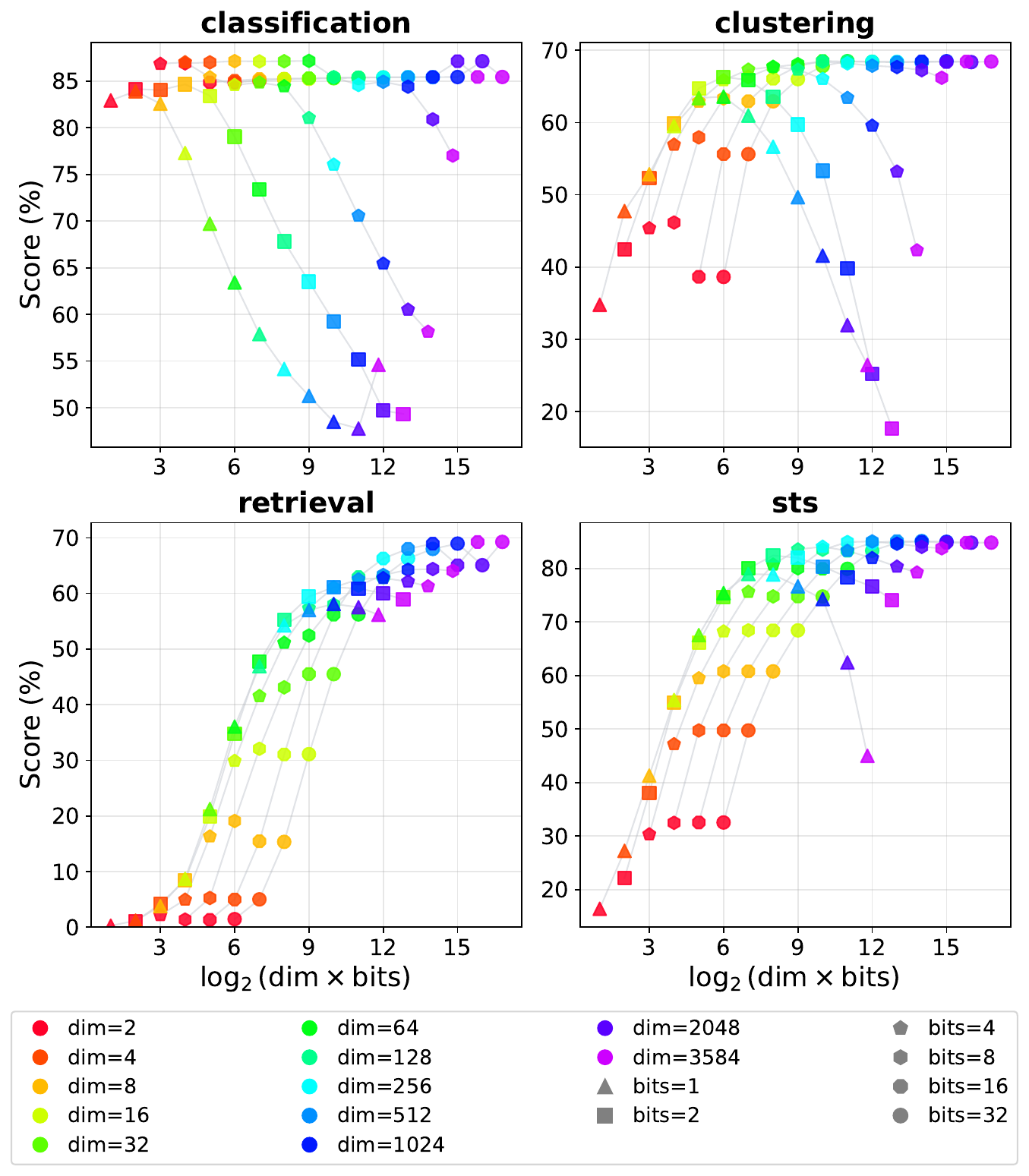}
\caption{
Performance of PCA without random orthogonal rotation for gte-Qwen2 across
different task types, bit-widths, and target dimensions.}
\label{fig:pca_gte-Qwen2}
\end{figure}

We further examine the role of random orthogonal rotation after PCA. PCA alone
produces coordinates ordered by explained variance, which can interact poorly
with scalar quantization because coordinate scales are highly imbalanced.
Figure~\ref{fig:pca_gte-Qwen2} shows that, for gte-Qwen2, increasing
the number of retained PCA dimensions does not always improve performance at a
fixed bit-width. This non-monotonic behavior is counterintuitive because adding
dimensions should, in principle, preserve more information before
quantization. We hypothesize that additional low-variance components can be
strongly affected by discretization noise, while high-variance components
dominate the geometry of the embedding space. As a result, adding more PCA
dimensions can introduce noisy or unstable coordinates rather than consistently
improving downstream performance.

Random orthogonal rotation redistributes variance across coordinates within the PCA subspace while preserving inner products. 
The comparison between PCA and PCA+ROR suggests that this redistribution can improve the compatibility of PCA-based dimensionality reduction with low-bit scalar quantization.

Because the orthogonal rotation is randomized, we also evaluate the sensitivity of PCA+ROR to its random seed using gte-Qwen2. 
The three-seed comparison in Appendix~\ref{app:rotation_seed} shows that scores are nearly unchanged at most dimensionalities.
Seed dependence is more visible for classification at small dimensions; STS also varies in the most aggressive 1-bit, two-dimensional setting.

\section{Conclusion and Future Work}
\label{sec:conclusion}
We investigated how dimensionality reduction and quantization jointly affect
text embedding compression. Across four MTEB task families, we found that
these two compression axes are complementary: many
model--task settings can retain near-original performance with only a small
fraction of the original storage footprint. The best compression strategy,
however, depends strongly on the task and model. Classification and clustering
often tolerate aggressive compression, whereas retrieval generally requires a
larger retained subspace to preserve fine-grained similarity structure.

Our analysis further shows that the choice of dimensionality reduction and
quantization format matters. PCA followed by random orthogonal rotation is
effective under aggressive compression, while Head-based truncation achieves
higher retrieval scores for four of the five models at a 10\% storage budget.
We also find that fixed low-bit floating-point formats can behave
unexpectedly for normalized high-dimensional embeddings, highlighting the
importance of matching the quantizer to the embedding distribution.

Future work should investigate why configurations with the same storage footprint
but different dimension--bit-width balances yield different downstream
performance. For example, \citet{inkiriwang-etal-2025-really} proposed an intrinsic
evaluation framework that uses geometric and spectral measures to assess
compressed embeddings and relate them to downstream performance. Applying
such measures to our joint compression settings, while separately examining
the effects of discarded dimensions and quantization error, could help
explain when retaining more dimensions is more valuable than increasing
precision per coordinate and why this balance varies across tasks.

\section*{Limitations}

This study has two main limitations. First, we evaluate only a limited set of
compression methods. For dimensionality reduction, we focus on Head-based
truncation and PCA followed by random orthogonal rotation, and for quantization, we use
simple post-training scalar quantization with a fixed set of bit-widths. Thus,
our experiments do not exhaustively cover the design space of embedding
compression. Other methods, such as learned projections, product quantization,
vector quantization, or adaptive quantization, may produce different
compression--performance trade-offs.

Second, our evaluation is limited to a subset of MTEB. Although we consider four
major task families, classification, clustering, retrieval, and STS, the number
of datasets is limited and all experiments are conducted in English. Therefore,
it remains unclear whether the observed task- and model-dependent trends
generalize to other languages, domains, or task types. Future work should
validate these findings on larger and more diverse benchmarks, including
multilingual and domain-specific datasets.
\bibliography{custom}

\appendix

\label{sec:appendix}

\section{Model Details Used in Our Experiments}
Table~\ref{tab:model_details} summarizes the pretrained text embedding models
used in our experiments.
\label{app:models}
\begin{table*}[t]
\centering
\small
\begin{tabularx}{\textwidth}{@{}l>{\raggedright\arraybackslash}Xllrrc@{}}
\toprule
\textbf{Model} & \textbf{HuggingFace} & \textbf{License} & \textbf{Input format} & \textbf{Dim.} & \textbf{\#Params} & \textbf{MRL} \\
\midrule
gte-Qwen2
& \href{https://huggingface.co/Alibaba-NLP/gte-Qwen2-7B-instruct}{Alibaba-NLP/gte-Qwen2-7B-instruct}
& Apache-2.0 & Instruction & 3584 & 7.61B & No \\

Qwen3-Embedding
& \href{https://huggingface.co/Qwen/Qwen3-Embedding-8B}{Qwen/Qwen3-Embedding-8B}
& Apache-2.0 & Instruction & 4096 & 8B & Yes \\

E5-mistral
& \href{https://huggingface.co/intfloat/E5-mistral-7b-instruct}{intfloat/E5-mistral-7b-instruct}
& MIT & Instruction & 4096 & 7.11B & No \\

E5-large
& \href{https://huggingface.co/intfloat/e5-large-v2}{intfloat/e5-large-v2}
& MIT & Prefix & 1024 & 335M & No \\

Nomic Embed v1.5
& \href{https://huggingface.co/nomic-ai/nomic-embed-text-v1.5}{nomic-ai/nomic-embed-text-v1.5}
& Apache-2.0 & Prefix & 768 & 137M & Yes \\

\bottomrule
\end{tabularx}
\caption{Details of the text embedding models used in our experiments. License identifiers follow the corresponding Hugging Face model cards.}
\label{tab:model_details}
\end{table*}

For the instruction-based models in Table~\ref{tab:model_details}, we encode
input texts with task-specific instructions for all tasks. For classification,
clustering, and STS tasks, we use the following general prompt format:
\[
    \texttt{Instruct: \{instruction\}\textbackslash nInput: \{text\}}.
\]
For retrieval tasks, we encode queries as
\[
    \texttt{Instruct: \{instruction\}\textbackslash nQuery: \{text\}},
\]
whereas corpus documents are encoded without an instruction prompt, following
the standard asymmetric retrieval setting.

For E5-large, we use its
standard input format. Specifically, query-side texts are encoded with the
\texttt{query:} prefix, and corpus or document-side texts are encoded with the
\texttt{passage:} prefix. Thus, for retrieval tasks, queries are encoded as
\texttt{query: \{text\}}, while corpus documents are encoded as
\texttt{passage: \{text\}}. For classification, clustering, and STS tasks,
where there is no explicit corpus/document side, we encode each input text as a
query-side text using the \texttt{query:} prefix.

For nomic-embed-text-v1.5, retrieval queries use the
\texttt{search\_query:} prefix and corpus documents use the
\texttt{search\_document:} prefix. Classification and clustering inputs use
the \texttt{classification:} and \texttt{clustering:} prefixes,
respectively. Following the evaluation setting of prior work~\cite{tsukagoshi-sasano-2025-redundancy},
we encode STS inputs without a prefix.

\section{Dataset-Specific Instructions}
\label{app:instructions}
Table~\ref{tab:dataset_instructions} lists the dataset-specific instructions
used for instruction-based embedding models. These instructions are prepended
to input texts when encoding classification, clustering, STS, and retrieval
queries. For retrieval tasks, corpus documents are encoded without these
instructions, following the asymmetric retrieval setting.

\begin{table*}[t]
\centering
\small
\begin{tabularx}{\textwidth}{@{}p{0.30\textwidth}X@{}}
\toprule
\textbf{Task} & \textbf{Instruction} \\
\midrule
AmazonCounterfactualClassification
& Classify a given Amazon customer review text as either counterfactual or not-counterfactual \\
AmazonPolarityClassification
& Classify Amazon reviews into positive or negative sentiment \\
AmazonReviewsClassification
& Classify the given Amazon review into its appropriate rating category \\
ImdbClassification
& Classify the sentiment expressed in the given movie review text from the IMDB dataset \\
ToxicConversationsClassification
& Classify the given comments as either toxic or not toxic \\
\midrule
ArxivClusteringS2S
& Identify the main and secondary category of Arxiv papers based on the titles \\
RedditClustering
& Identify the topic or theme of Reddit posts based on the titles \\
StackExchangeClustering
& Identify the topic or theme of StackExchange posts based on the titles \\
\midrule
MIRACLRetrievalHardNegatives
& Given a question, retrieve Wikipedia passages that answer the question \\
QuoraRetrievalHardNegatives
& Given a question, retrieve questions that are semantically equivalent to the given question \\
HotpotQAHardNegatives
& Given a multi-hop question, retrieve documents that can help answer the question \\
DBPediaHardNegatives
& Given a query, retrieve relevant entity descriptions from DBPedia \\
NQHardNegatives
& Given a question, retrieve Wikipedia passages that answer the question \\
MSMARCOHardNegatives
& Given a web search query, retrieve relevant passages that answer the query \\
\midrule
STS-12
& Retrieve semantically similar text \\
STS-13
& Retrieve semantically similar text \\
STS-14
& Retrieve semantically similar text \\
STS-15
& Retrieve semantically similar text \\
STS-16
& Retrieve semantically similar text \\
STS-17
& Retrieve semantically similar text \\
STS-Benchmark
& Retrieve semantically similar text \\
SICK-R
& Retrieve semantically similar text \\
\bottomrule
\end{tabularx}
\caption{Dataset-specific instructions used for instruction-based embedding models.}
\label{tab:dataset_instructions}
\end{table*}

\section{Fixed-Budget Performance}
\label{app:budget_performance}
Table~\ref{tab:budget_performance} reports the best observed performance
within each relative embedding storage budget for all five models. For each
task family and reduction method, the table gives the highest score among
evaluated dimensionality and bit-width combinations that satisfy the budget,
along with the selected combination. Scores are expressed relative to the
original full-dimensional 32-bit Head embeddings.

Across all five models, classification remains robust even at a 0.1\%
storage budget: each model retains at least 96.5\% of its original score.
Retrieval requires substantially more storage. The three high-dimensional
models recover approximately 94--96\% of their original retrieval scores
within a 1\% budget, whereas E5-large and Nomic reach comparable performance
at a 10\% budget. This pattern is consistent with retrieval requiring a
minimum amount of retained information, reached at different relative
budgets across models.

MRL compatibility does not yield a consistent advantage under Head-based
compression. Although the MRL-compatible Qwen3-Embedding performs well,
the other MRL-compatible model, Nomic, is often less robust than the non-MRL
gte-Qwen2. Thus, MRL status alone does not predict compression robustness
across these models and tasks.

\begin{table*}[t!]
\centering
\footnotesize
\setlength{\tabcolsep}{3pt}
\renewcommand{\arraystretch}{0.8}
\begin{tabular}{@{}lll r@{\hspace{0.4em}}l r@{\hspace{0.4em}}l r@{\hspace{0.4em}}l r@{\hspace{0.4em}}l@{}}
\toprule
Model & Budget & Method & \multicolumn{2}{c}{Classification} & \multicolumn{2}{c}{Clustering} & \multicolumn{2}{c}{Retrieval} & \multicolumn{2}{c}{STS} \\
\midrule
\multirow{8}{*}{gte-Qwen2} & \multirow{2}{*}{0.01\%} & Head & 93.2 & ($8\times1$) & 57.1 & ($8\times1$) & 2.6 & ($8\times1$) & 53.5 & ($4\times2$) \\
 &  & PCA+ROR & 99.5 & ($2\times4$) & 77.6 & ($8\times1$) & 5.9 & ($8\times1$) & 49.1 & ($8\times1$) \\
\cmidrule{2-11}
 & \multirow{2}{*}{0.1\%} & Head & 97.8 & ($32\times2$) & 92.9 & ($32\times2$) & 48.0 & ($32\times2$) & 87.1 & ($32\times2$) \\
 &  & PCA+ROR & 100.0 & ($8\times8$) & 98.1 & ($32\times2$) & 52.1 & ($64\times1$) & 86.6 & ($64\times1$) \\
\cmidrule{2-11}
 & \multirow{2}{*}{1\%} & Head & 99.5 & ($32\times4$) & 99.3 & ($256\times4$) & 96.0 & ($512\times2$) & 99.1 & ($1024\times1$) \\
 &  & PCA+ROR & 100.0 & ($64\times16$) & 100.1 & ($128\times4$) & 95.3 & ($512\times2$) & 100.4 & ($512\times2$) \\
\cmidrule{2-11}
 & \multirow{2}{*}{10\%} & Head & 99.9 & ($2048\times4$) & 99.9 & ($512\times4$) & 99.2 & ($3584\times2$) & 100.4 & ($3584\times2$) \\
 &  & PCA+ROR & 100.0 & ($256\times16$) & 100.2 & ($256\times32$) & 98.9 & ($2048\times4$) & 100.5 & ($512\times8$) \\
\midrule
\multirow{8}{*}{Qwen3-Emb.} & \multirow{2}{*}{0.01\%} & Head & 92.1 & ($4\times2$) & 37.6 & ($4\times2$) & 2.0 & ($8\times1$) & 54.3 & ($4\times2$) \\
 &  & PCA+ROR & 99.4 & ($2\times4$) & 66.7 & ($8\times1$) & 6.6 & ($4\times2$) & 51.8 & ($8\times1$) \\
\cmidrule{2-11}
 & \multirow{2}{*}{0.1\%} & Head & 99.0 & ($64\times2$) & 92.9 & ($64\times2$) & 70.7 & ($64\times2$) & 94.6 & ($32\times4$) \\
 &  & PCA+ROR & 100.1 & ($8\times8$) & 99.3 & ($64\times2$) & 71.6 & ($128\times1$) & 93.2 & ($128\times1$) \\
\cmidrule{2-11}
 & \multirow{2}{*}{1\%} & Head & 99.7 & ($512\times2$) & 99.0 & ($512\times2$) & 95.1 & ($512\times2$) & 99.3 & ($256\times4$) \\
 &  & PCA+ROR & 100.1 & ($8\times8$) & 100.7 & ($128\times8$) & 95.9 & ($512\times2$) & 98.6 & ($512\times2$) \\
\cmidrule{2-11}
 & \multirow{2}{*}{10\%} & Head & 100.0 & ($4096\times2$) & 100.4 & ($2048\times4$) & 99.1 & ($3584\times2$) & 100.0 & ($1024\times8$) \\
 &  & PCA+ROR & 100.1 & ($8\times8$) & 100.7 & ($128\times8$) & 98.9 & ($4096\times2$) & 98.9 & ($1024\times4$) \\
\midrule
\multirow{8}{*}{E5-mistral} & \multirow{2}{*}{0.01\%} & Head & 96.2 & ($4\times2$) & 41.1 & ($4\times2$) & 1.0 & ($8\times1$) & 53.5 & ($4\times2$) \\
 &  & PCA+ROR & 98.4 & ($2\times4$) & 68.7 & ($4\times2$) & 6.7 & ($8\times1$) & 49.7 & ($8\times1$) \\
\cmidrule{2-11}
 & \multirow{2}{*}{0.1\%} & Head & 100.8 & ($16\times8$) & 88.3 & ($128\times1$) & 51.6 & ($64\times2$) & 90.6 & ($32\times4$) \\
 &  & PCA+ROR & 100.0 & ($16\times8$) & 99.1 & ($64\times2$) & 63.8 & ($64\times2$) & 91.7 & ($128\times1$) \\
\cmidrule{2-11}
 & \multirow{2}{*}{1\%} & Head & 100.8 & ($16\times8$) & 99.2 & ($512\times2$) & 91.8 & ($512\times2$) & 99.4 & ($512\times2$) \\
 &  & PCA+ROR & 100.6 & ($1024\times1$) & 100.8 & ($512\times2$) & 94.4 & ($512\times2$) & 99.0 & ($512\times2$) \\
\cmidrule{2-11}
 & \multirow{2}{*}{10\%} & Head & 100.8 & ($16\times8$) & 100.3 & ($2048\times4$) & 99.5 & ($4096\times2$) & 100.5 & ($4096\times2$) \\
 &  & PCA+ROR & 105.5 & ($3584\times1$) & 101.1 & ($3584\times2$) & 99.7 & ($2048\times4$) & 99.7 & ($1024\times8$) \\
\midrule
\multirow{8}{*}{E5-large} & \multirow{2}{*}{0.01\%} & Head & 62.0 & ($2\times1$) & 5.6 & ($2\times1$) & 0.1 & ($2\times1$) & 23.5 & ($2\times1$) \\
 &  & PCA+ROR & 90.4 & ($2\times1$) & 38.8 & ($2\times1$) & 0.6 & ($2\times1$) & 14.6 & ($2\times1$) \\
\cmidrule{2-11}
 & \multirow{2}{*}{0.1\%} & Head & 76.0 & ($8\times4$) & 20.5 & ($16\times2$) & 9.2 & ($16\times2$) & 75.5 & ($8\times4$) \\
 &  & PCA+ROR & 99.4 & ($8\times4$) & 69.6 & ($16\times2$) & 21.6 & ($32\times1$) & 69.2 & ($32\times1$) \\
\cmidrule{2-11}
 & \multirow{2}{*}{1\%} & Head & 91.0 & ($128\times2$) & 70.8 & ($128\times2$) & 58.2 & ($128\times2$) & 96.6 & ($64\times4$) \\
 &  & PCA+ROR & 100.1 & ($64\times4$) & 96.5 & ($128\times2$) & 69.5 & ($128\times2$) & 96.3 & ($256\times1$) \\
\cmidrule{2-11}
 & \multirow{2}{*}{10\%} & Head & 99.4 & ($1024\times2$) & 97.2 & ($1024\times2$) & 95.9 & ($1024\times2$) & 100.4 & ($256\times8$) \\
 &  & PCA+ROR & 100.2 & ($64\times8$) & 100.0 & ($512\times4$) & 95.4 & ($512\times4$) & 100.4 & ($1024\times2$) \\
\midrule
\multirow{8}{*}{Nomic} & \multirow{2}{*}{0.01\%} & Head & 63.5 & ($2\times1$) & 6.9 & ($2\times1$) & 0.4 & ($2\times1$) & 23.6 & ($2\times1$) \\
 &  & PCA+ROR & 88.7 & ($2\times1$) & 46.6 & ($2\times1$) & 0.2 & ($2\times1$) & 18.0 & ($2\times1$) \\
\cmidrule{2-11}
 & \multirow{2}{*}{0.1\%} & Head & 69.7 & ($8\times2$) & 28.3 & ($16\times1$) & 5.5 & ($16\times1$) & 62.3 & ($4\times4$) \\
 &  & PCA+ROR & 96.5 & ($4\times4$) & 69.6 & ($8\times2$) & 12.0 & ($16\times1$) & 62.7 & ($16\times1$) \\
\cmidrule{2-11}
 & \multirow{2}{*}{1\%} & Head & 86.4 & ($64\times2$) & 82.2 & ($64\times2$) & 66.0 & ($64\times2$) & 93.9 & ($64\times2$) \\
 &  & PCA+ROR & 100.2 & ($16\times8$) & 95.2 & ($64\times2$) & 63.4 & ($128\times1$) & 93.8 & ($64\times2$) \\
\cmidrule{2-11}
 & \multirow{2}{*}{10\%} & Head & 99.5 & ($512\times4$) & 99.2 & ($512\times4$) & 98.9 & ($512\times4$) & 99.8 & ($256\times8$) \\
 &  & PCA+ROR & 100.2 & ($16\times8$) & 100.1 & ($256\times8$) & 97.7 & ($512\times4$) & 100.4 & ($512\times2$) \\
\bottomrule
\end{tabular}
\caption{Best observed performance within fixed embedding storage budgets.
Each score is the highest evaluated task-family mean, expressed as a percentage
of the original full-dimensional 32-bit Head score, among configurations satisfying
$100 d b / (32 D) \leq \text{Budget}$, where $D$ is the original dimensionality.
Parentheses give the dimensionality and bit-width $(d\times b)$ of the selected
configuration. Values above 100\% exceed the reference score.}
\label{tab:budget_performance}
\end{table*}

\section{Experimental Results for Other Models}
\label{app:all_results}

\begin{figure}
    \centering
    \includegraphics[width=\linewidth]{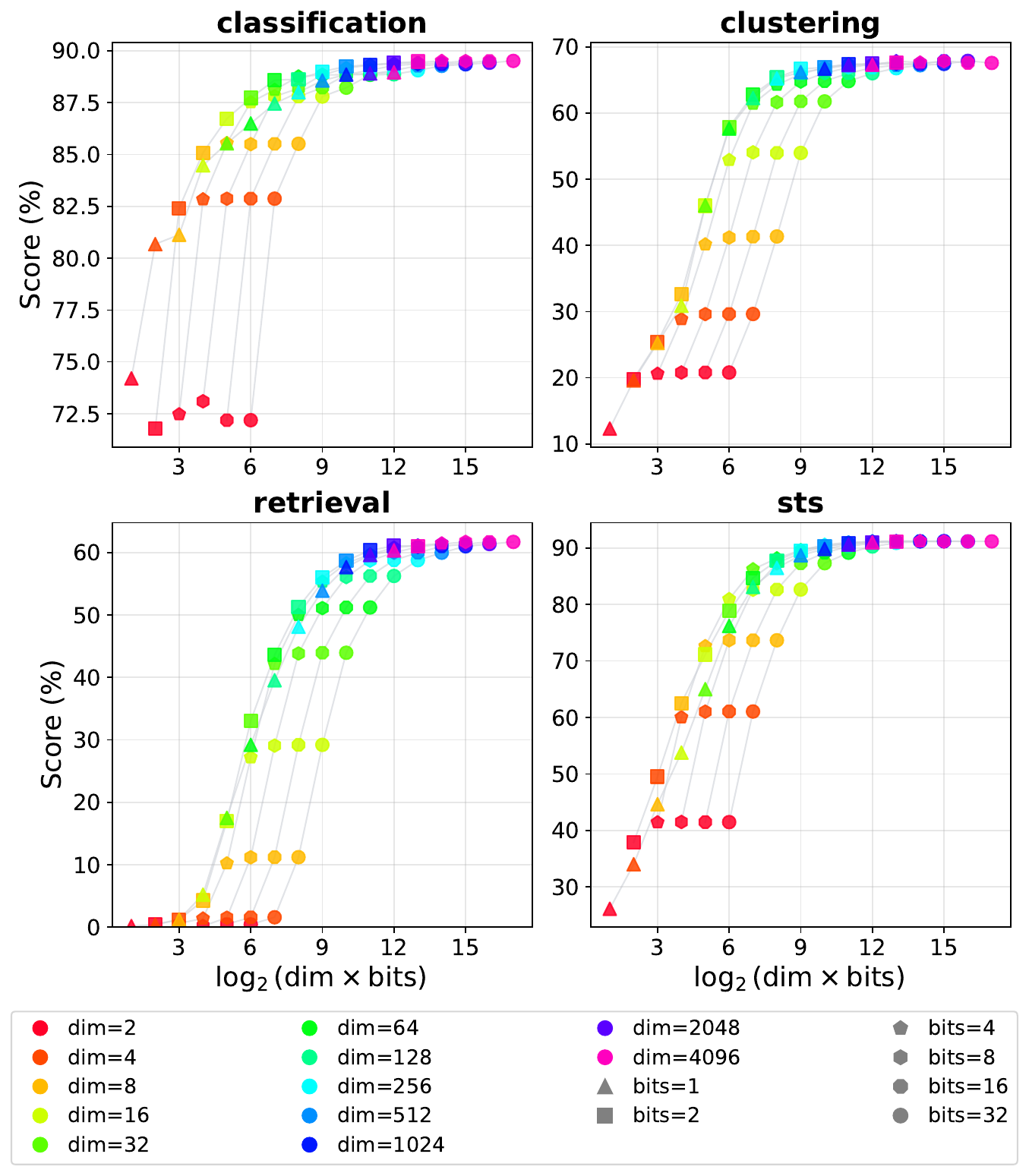}
    \includegraphics[width=\linewidth]{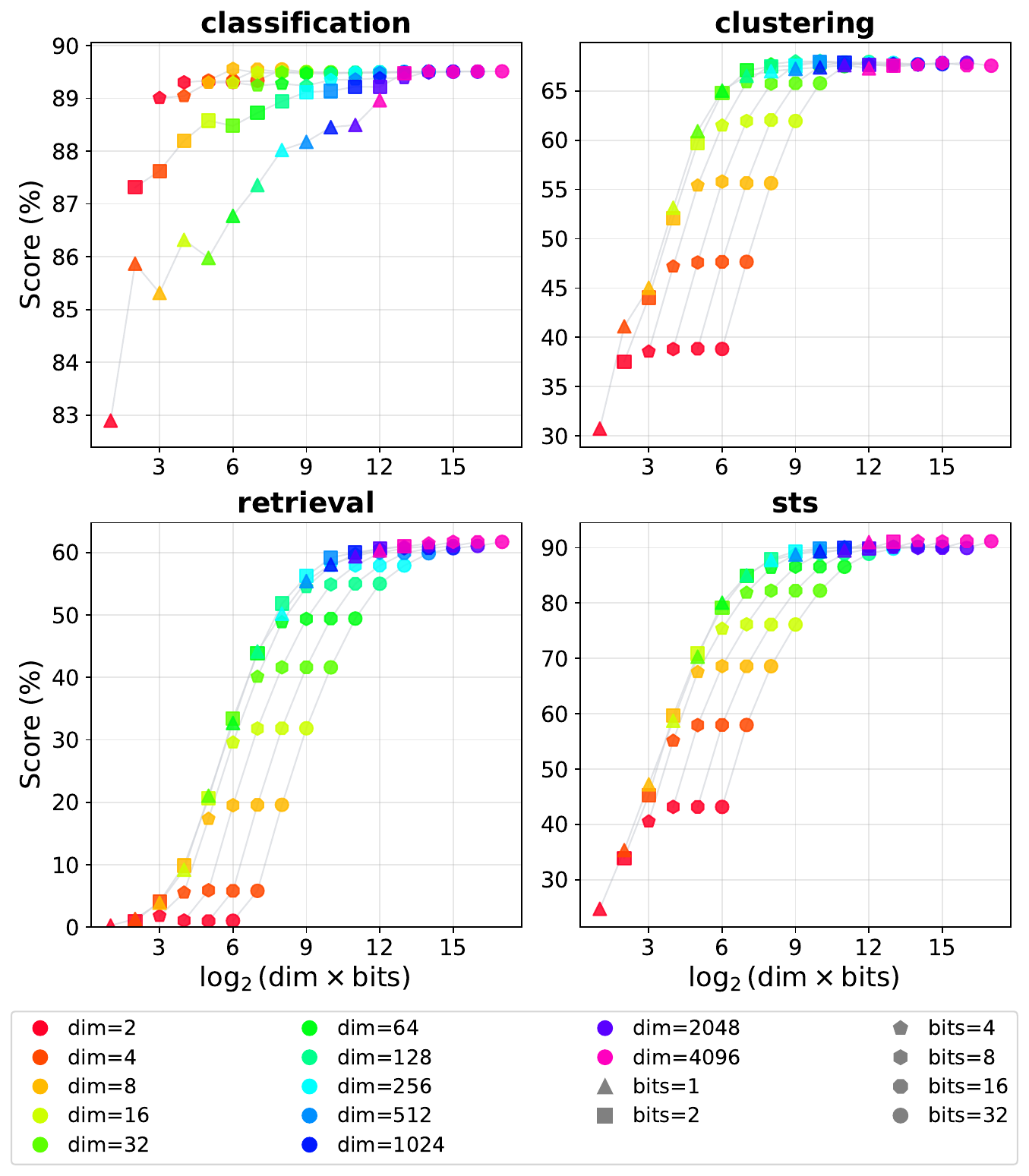}
    \caption{Performance of Head-based dimensionality reduction (top) and PCA+ROR (bottom) for Qwen3-Embedding across different bit-widths and task types.}
    \label{fig:qwen3_results}
\end{figure}

\begin{figure}
    \centering
    \includegraphics[width=\linewidth]{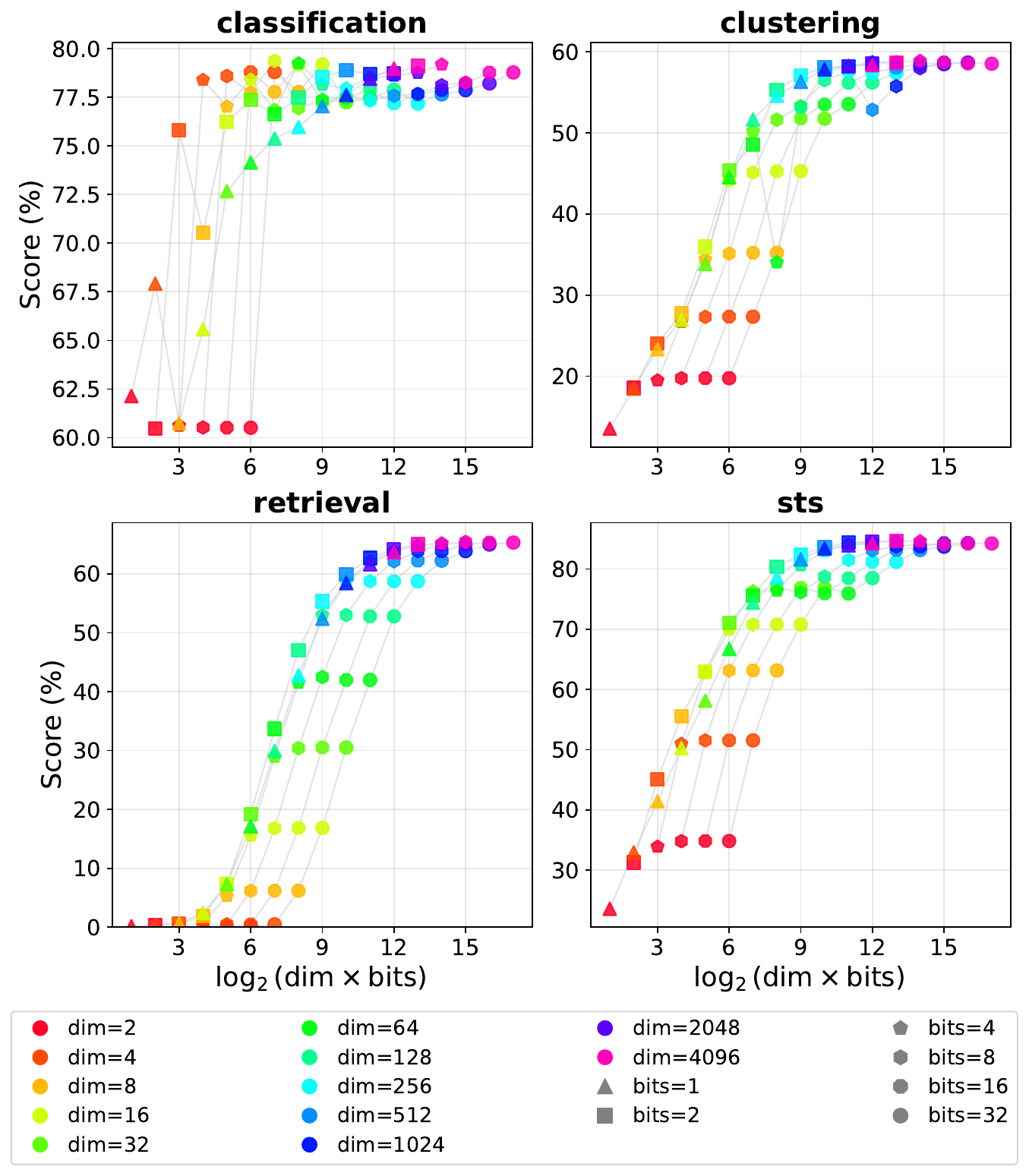}
    \includegraphics[width=\linewidth]{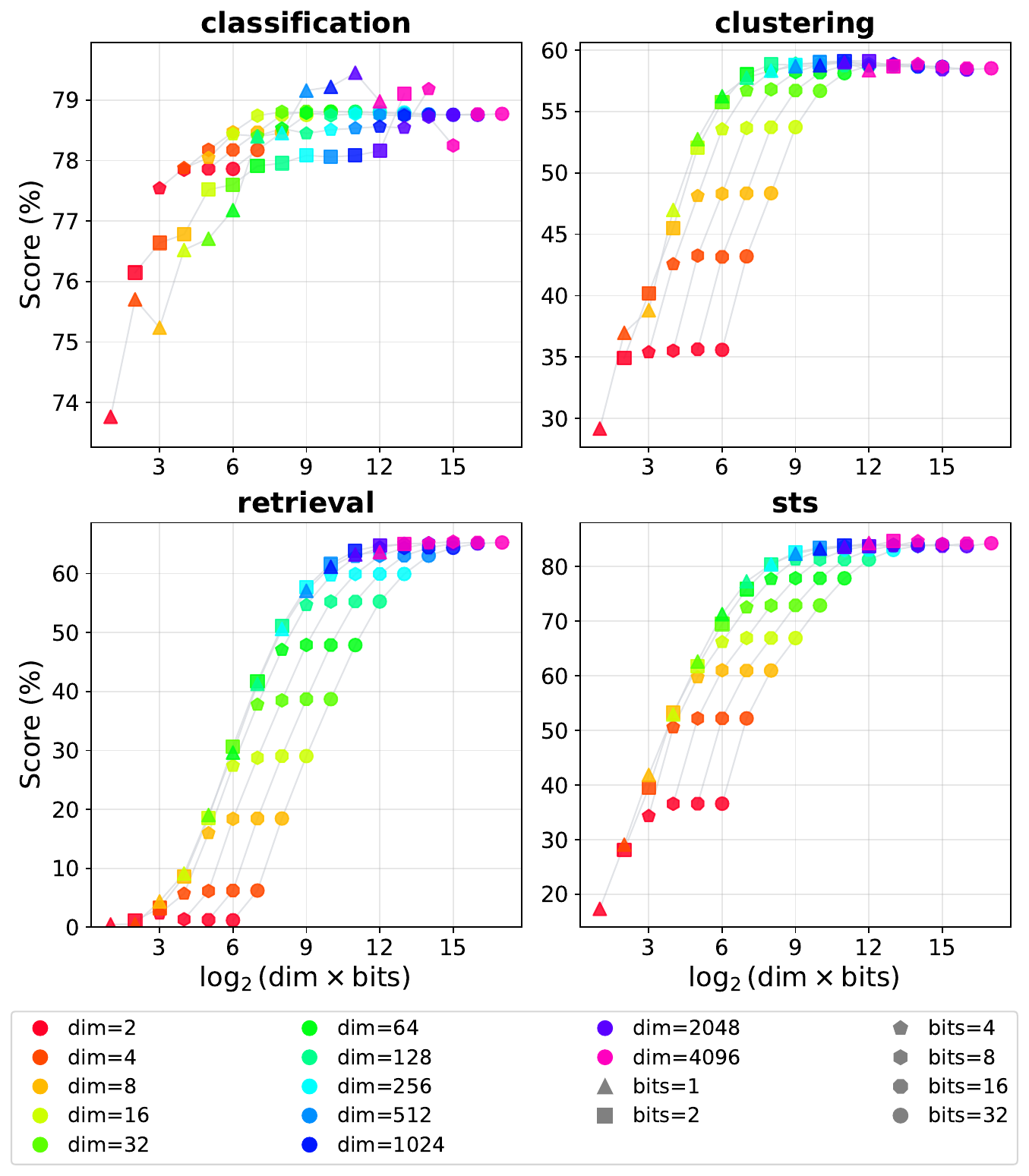}
    \caption{Performance of Head-based dimensionality reduction (top) and PCA+ROR (bottom) for E5-mistral across different bit-widths and task types.}
    \label{fig:e5_mistral_results}
\end{figure}

\begin{figure}
    \centering
    \includegraphics[width=\linewidth]{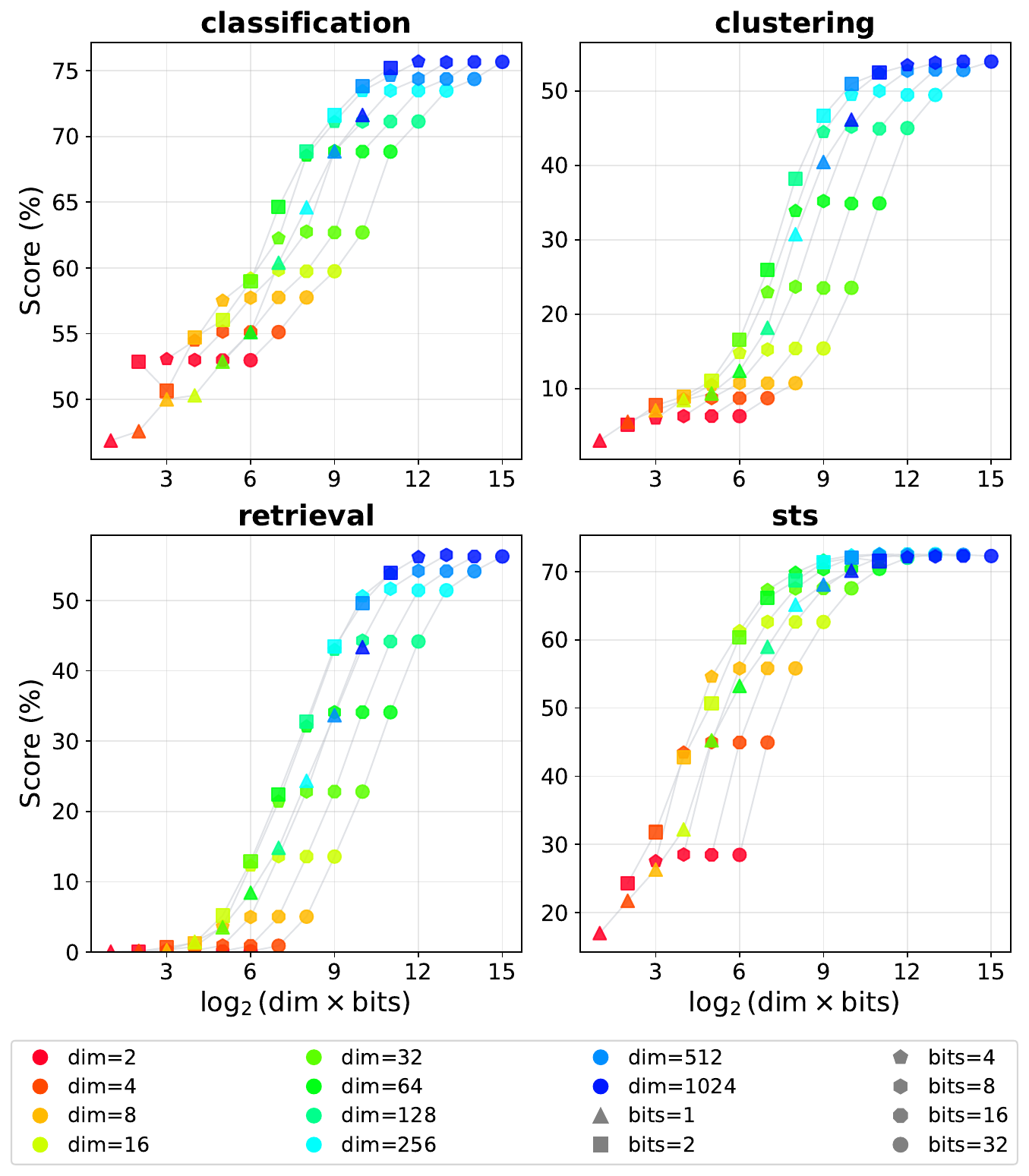}
    \includegraphics[width=\linewidth]{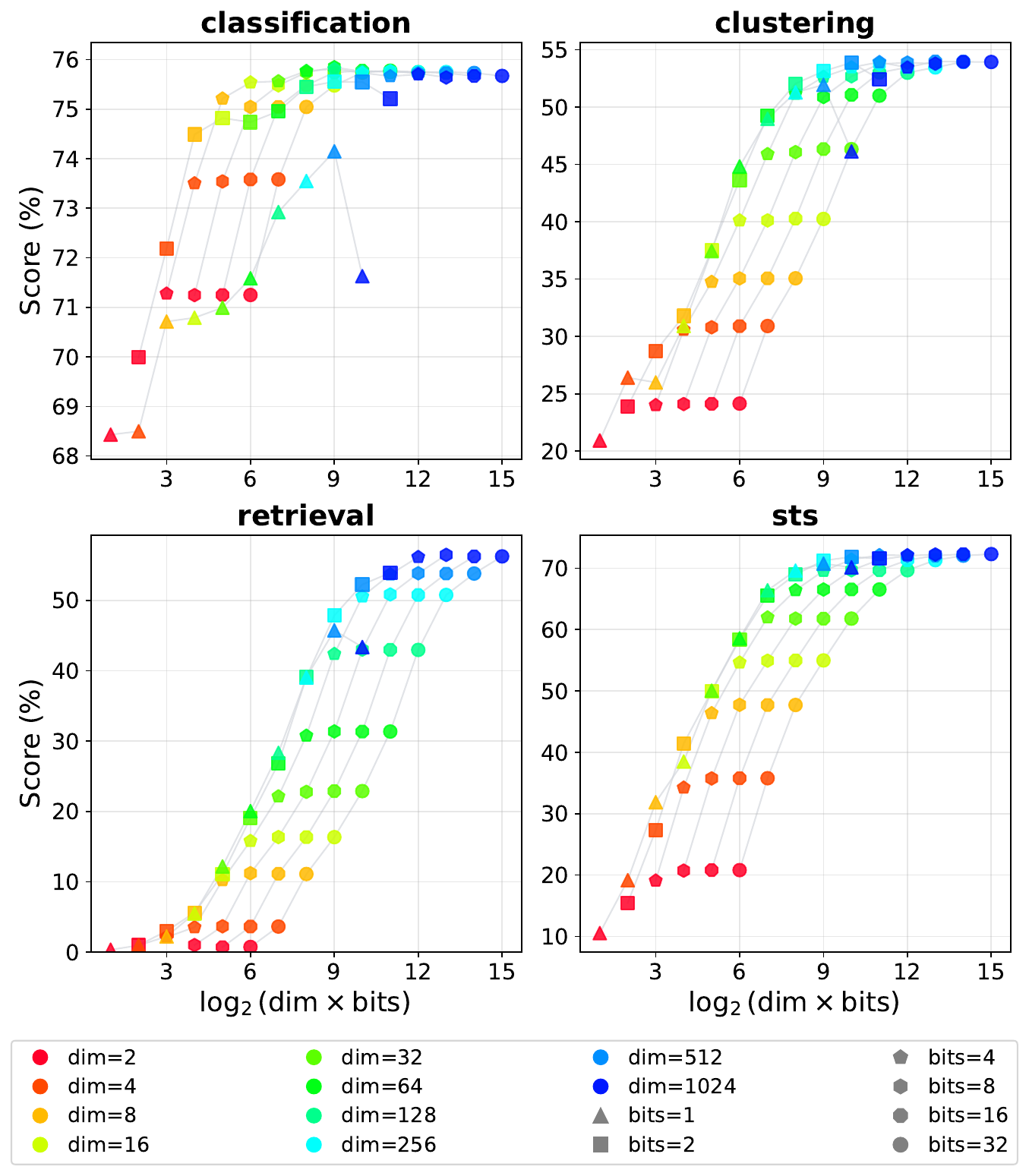}
    \caption{Performance of Head-based dimensionality reduction (top) and PCA+ROR (bottom) for E5-large across different bit-widths and task types.}
    \label{fig:e5_large_results}
\end{figure}

\begin{figure}
    \centering
    \includegraphics[width=\linewidth]{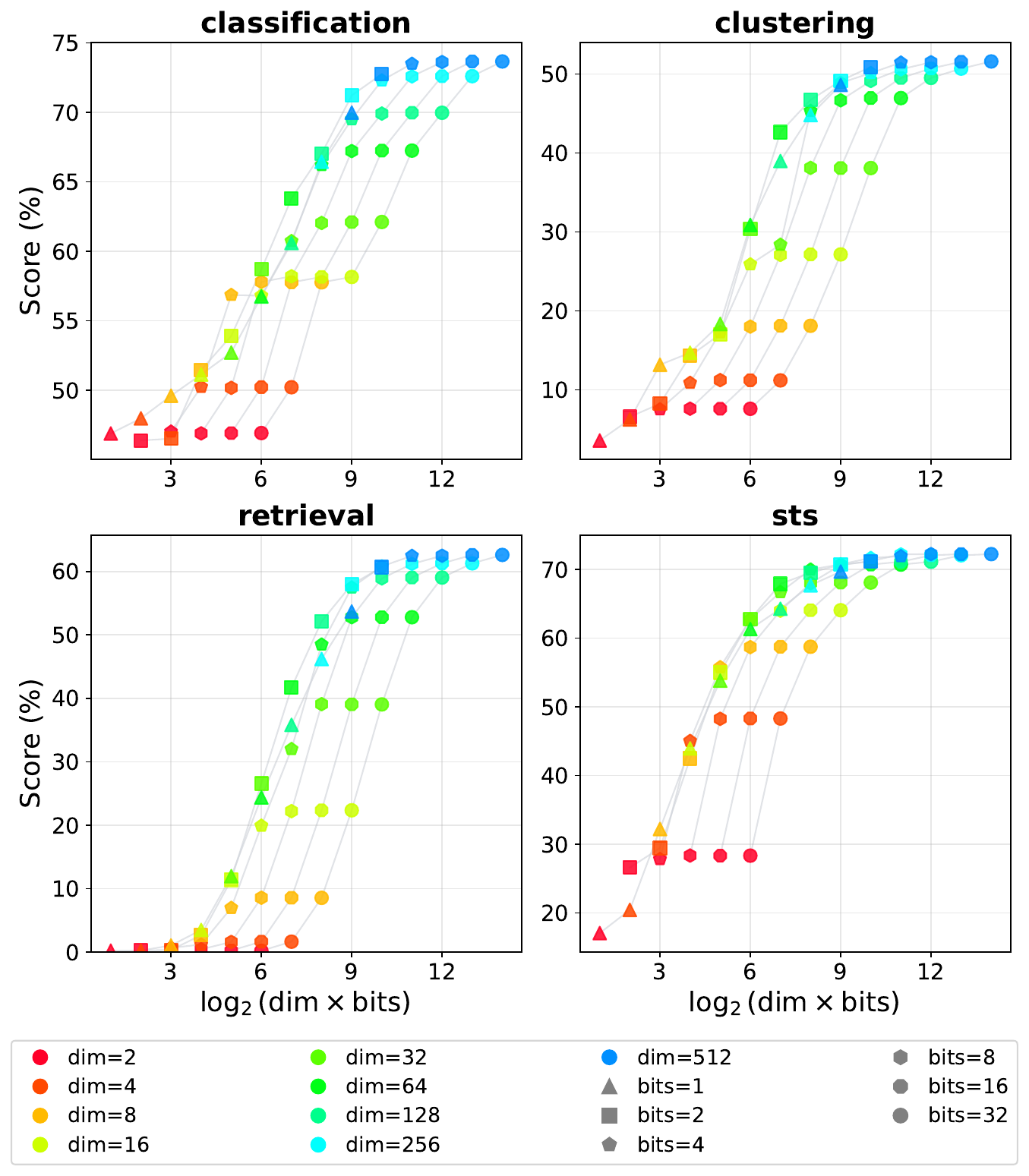}
    \includegraphics[width=\linewidth]{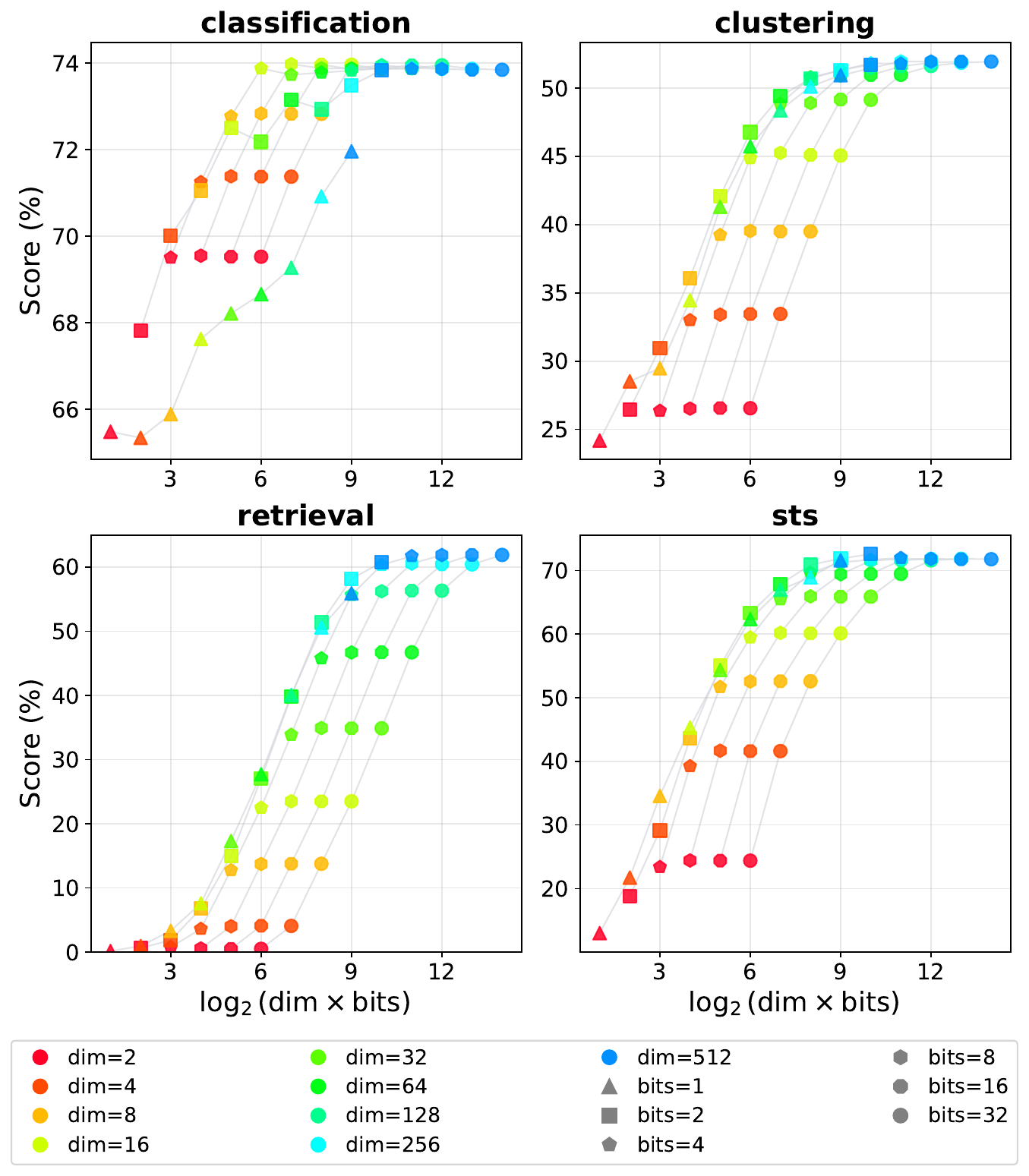}
    \caption{Performance of Head-based dimensionality reduction (top) and PCA+ROR (bottom) for Nomic Embed v1.5 across different bit-widths and task types.}
    \label{fig:nomic_results}
\end{figure}

Figures~\ref{fig:qwen3_results}, \ref{fig:e5_mistral_results},
\ref{fig:e5_large_results}, and~\ref{fig:nomic_results} show the performance
of Head-based dimensionality reduction and PCA+ROR for Qwen3-Embedding,
E5-mistral, E5-large, and Nomic Embed v1.5, respectively, across different
bit-widths and task types. These figures complement the results for gte-Qwen2
presented in the main text and cover all five evaluated models.

\section{Sensitivity to Quantization Format}
\label{app:quant_format}
\begin{figure}[htbp]
    \centering
    \includegraphics[width=\linewidth]{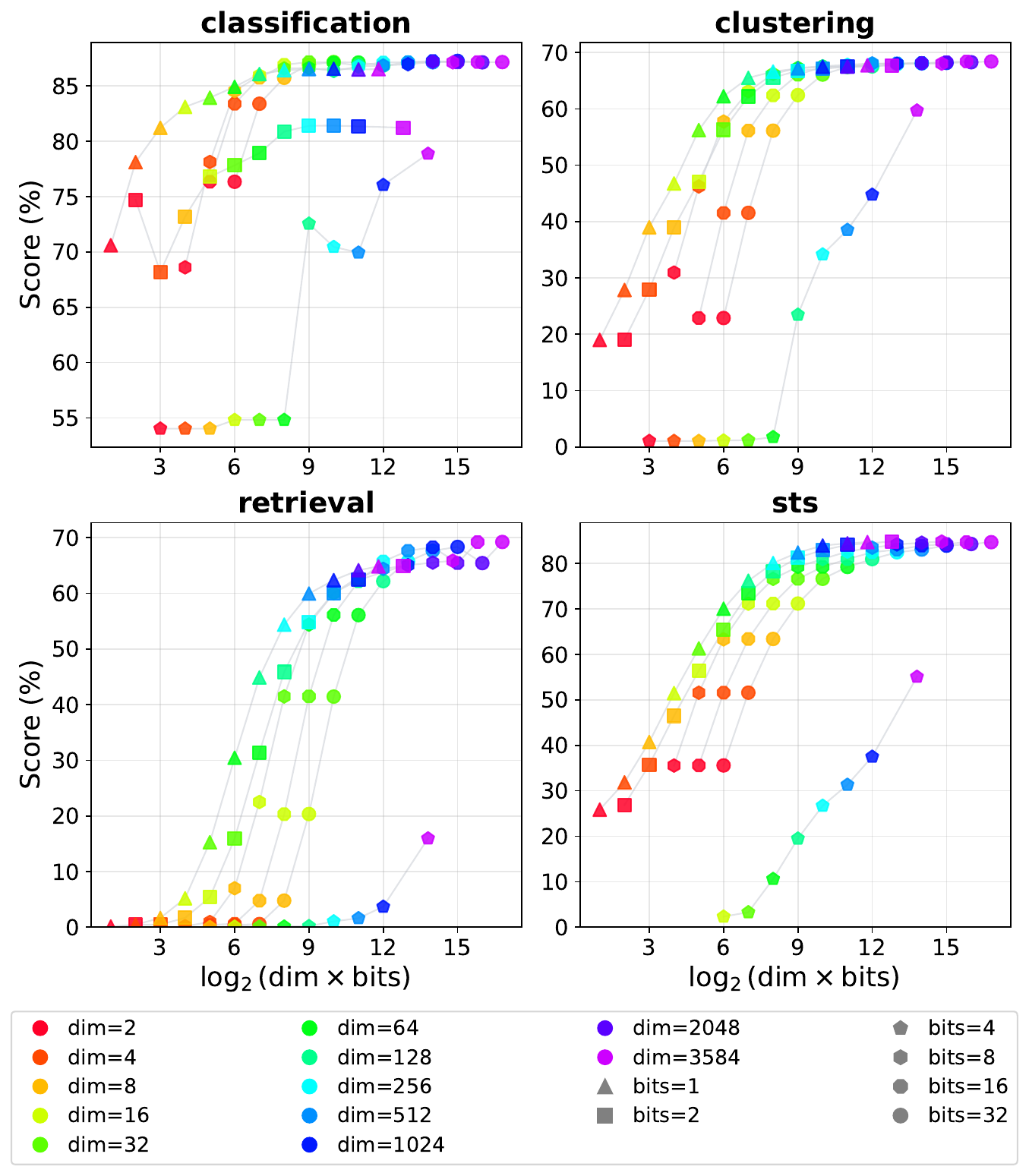}
\caption{
Effect of fixed quantization formats on gte-Qwen2 with
Head-based dimensionality reduction. }
    \label{fig:qwen2_fixed_quant}
\end{figure}

Our main experiments use a global equal-count lookup-table quantizer for
2-, 4-, and 8-bit quantization, rather than applying low-precision
floating-point formats directly. This choice is motivated by the value
distribution of text embeddings: since the embedding models used in this
study output L2-normalized vectors, individual coordinates tend to have
small magnitudes and are highly concentrated around zero. We further
examine this design choice by evaluating fixed quantization formats on
gte-Qwen2 with Head-based dimensionality reduction.

For this analysis, we replace the equal-count quantizer with fixed
representable value sets. For 2-bit quantization, we use the codebook
\(\{-0.75, -0.25, 0.25, 0.75\}\). For 4-bit quantization, we use a
floating-point format with one sign bit, one exponent bit, two mantissa
bits, bias 2, subnormal values enabled, and no reserved special exponent.
For 8-bit quantization, we use a floating-point format with one sign bit,
four exponent bits, three mantissa bits, and bias 7.

Figure~\ref{fig:qwen2_fixed_quant} shows that the 4-bit floating-point
setting performs substantially worse than the neighboring 2-bit and 8-bit
settings across several task families. This result is counterintuitive if
bit-width alone is considered, but it is consistent with the scale of
L2-normalized high-dimensional embeddings. In the 4-bit format used here,
the representable values near zero are too coarse, so many small-magnitude
coordinates are rounded to zero. This zero collapse can remove directional
information from the embedding, which is especially harmful for
similarity-based evaluation. By contrast, the 2-bit codebook does not
contain zero and therefore preserves at least the sign of every coordinate,
while the 8-bit floating-point format has sufficiently finer resolution
around zero.
This result supports our use of equal-count quantization, which matches
representable levels to the empirical embedding distribution and avoids
excessive collapse of small but informative coordinates.

\section{Sensitivity to Fitting--Evaluation Overlap}
\label{app:evaluation_overlap}

The main experiments fit PCA and quantization tables using embeddings drawn
from task inputs and evaluate on the full selected MTEB split. Consequently,
some inputs used for fitting are also evaluated. To assess the sensitivity of
our results to this overlap, we conduct an additional experiment with
Qwen3-Embedding-8B on the eight STS and five classification tasks, using their
test splits.

For each Hugging Face subset with \(N\) evaluation examples, we generate a
random permutation with seed 42 and partition the examples into \(F\), \(C\),
and \(R\), where \(|F|=|R|=\lfloor N/5\rfloor\) and
\(|C|=N-2\lfloor N/5\rfloor\). We fit the compression parameters using \(F\),
then compare evaluation on \(C\cup R\), which excludes fitting examples, with
evaluation on \(F\cup C\), which includes them. The two evaluation sets have
equal size and share the same examples in \(C\). The examples in \(F\) make up
approximately one quarter of the latter evaluation set.

This experiment samples approximately 20\% of evaluation examples, whereas
the main experiments sample distinct cached embeddings. STS sentence pairs
are partitioned as indivisible examples; both sentences from pairs in \(F\)
are collected for fitting. For classification, we collect the input texts in
\(F\). After removing duplicate fitting texts, we use their embeddings to
fit PCA and the 2-, 4-, and 8-bit equal-count quantization tables. The
fitting set is not enlarged when fewer texts are available for PCA. Both
evaluation conditions use the same fitting texts, PCA parameters, random
rotations, and quantization tables. The embedding model is not fine-tuned,
and the standard MTEB classifier training procedure is unchanged.

\begin{figure*}[t]
    \centering
    \includegraphics[width=0.86\textwidth,trim=0bp 95bp 0bp 0bp,clip]{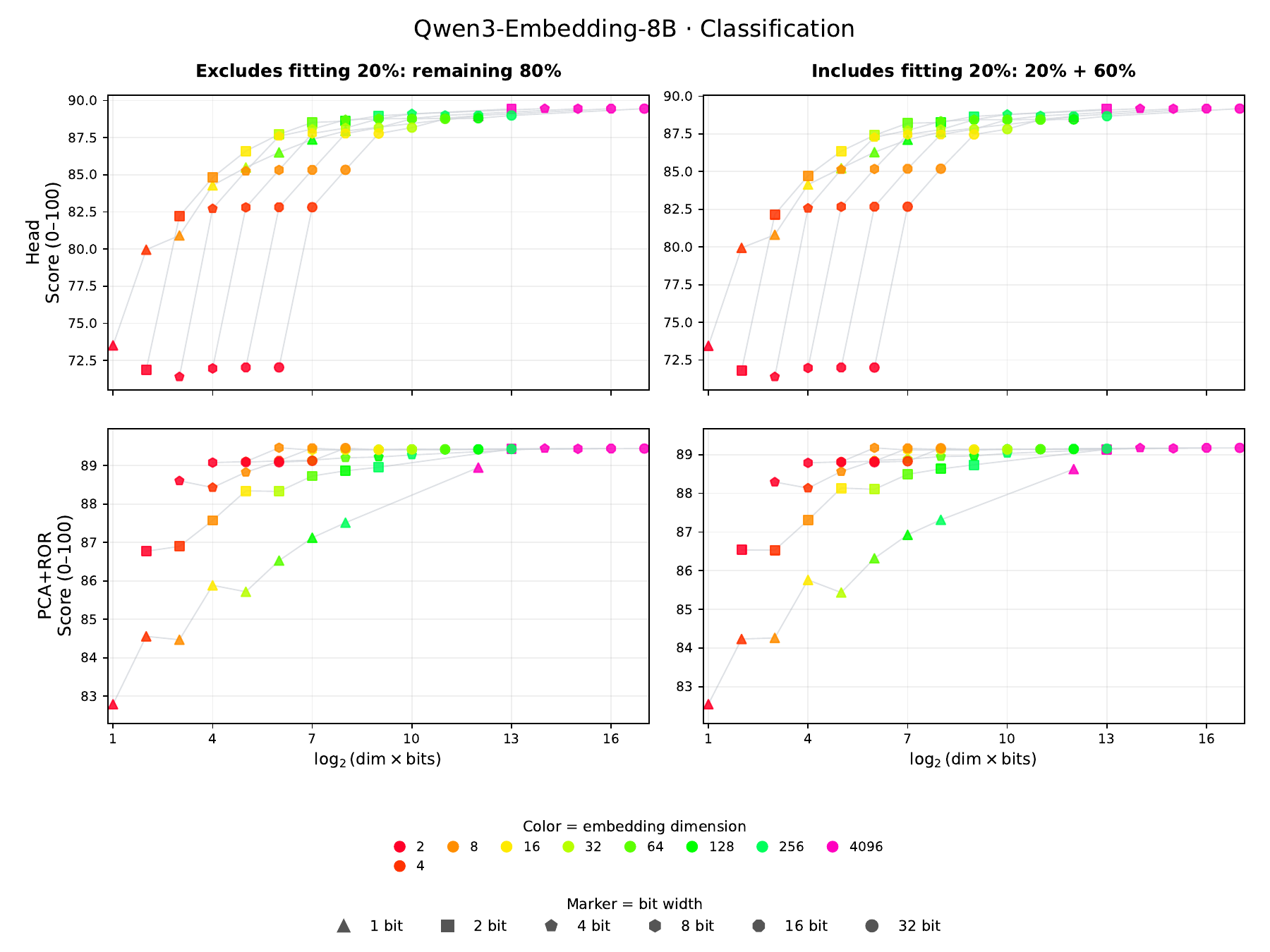}
    \par\vspace{0.5ex}
    \includegraphics[width=0.86\textwidth]{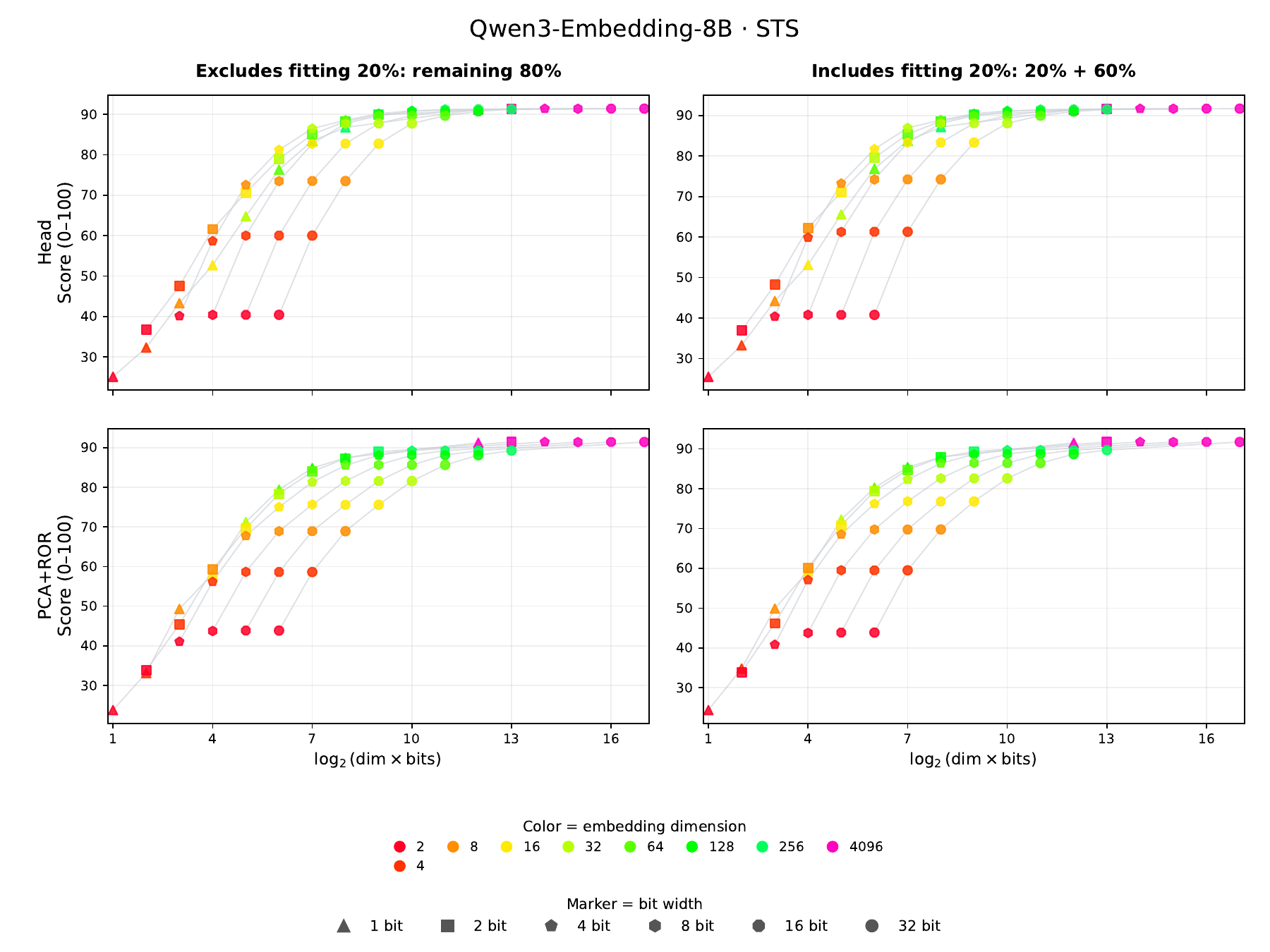}
    \caption{Effect of fitting--evaluation overlap for Qwen3-Embedding-8B on classification (upper graphic; five datasets) and STS (lower graphic; eight datasets). In each graphic, the left column excludes fitting examples and the right column includes them. The upper row shows Head and the lower row PCA+ROR. The legend below STS applies to both graphics.}
    \label{fig:overlap_results}
\end{figure*}

We compare dimensions \(d\in\{2,4,8,16,32,64,128,256\}\) for all 13 tasks,
so every plotted category average contains the same datasets at every
setting. Within each category, we average dataset main scores with equal
weight. Figure~\ref{fig:overlap_results}
shows that the aggregate scores are nearly unchanged when fitting examples
are included in the evaluation set, for both Head and PCA+ROR across the
plotted compression footprint. This suggests limited sensitivity to
fitting--evaluation overlap in this setting.
The 4096-dimensional points in the PCA+ROR panels are copied from Head as
full-dimensional references rather than measured PCA+ROR results. Vertical-axis
ranges differ between panels.

The split excludes fitting examples at the example level. Identical text may
still occur in different examples, particularly in STS sentence pairs.
Furthermore, the comparison replaces the examples in \(R\) with those in
\(F\), so a score difference cannot be attributed solely to input overlap.
The experiment uses one random split.

\section{Sensitivity to the Random Rotation Seed}
\label{app:rotation_seed}

PCA+ROR uses a random orthogonal matrix for each target dimension. To assess
the resulting variability, we repeat the gte-Qwen2 experiments with three
projection seeds, 1, 2, and 3, using the same model, tasks, fitting protocol,
target dimensions, and bit-widths. For each seed, we construct the rotation
using the projection seed plus the target dimension and fit the equal-count
quantization tables on the resulting rotated
calibration embeddings. We average dataset scores within each task family
and compare these task-family scores across seeds.

\begin{figure*}[t]
    \centering
    \includegraphics[width=0.77\textwidth,trim=0bp 30bp 785bp 55bp,clip]{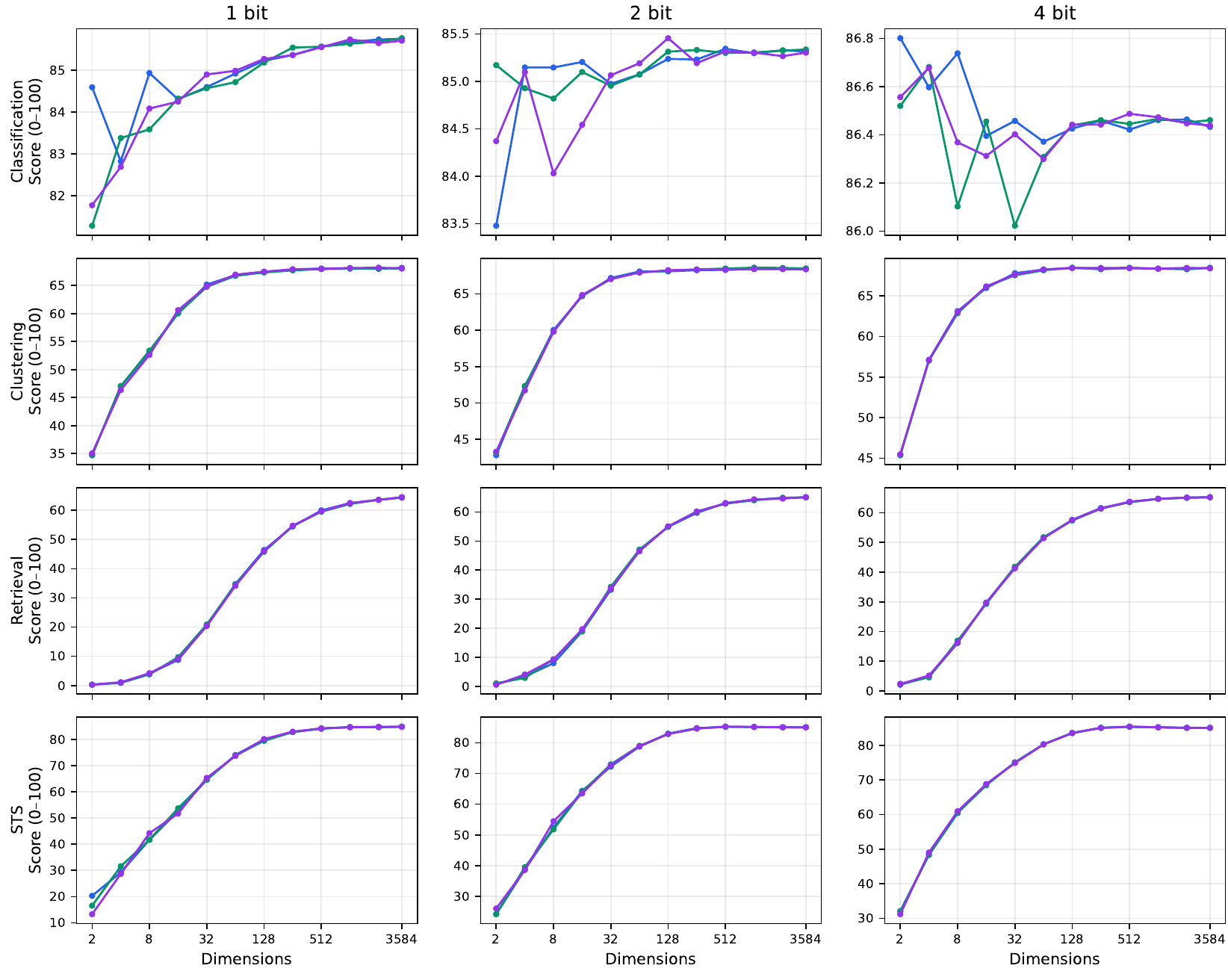}
    \par\vspace{0.5ex}
    \includegraphics[width=0.77\textwidth,trim=788bp 30bp 0bp 55bp,clip]{figure/random_seed/projection_seed_comparison.pdf}
    \caption{PCA+ROR scores for gte-Qwen2 across three random rotation seeds. The upper graphic shows 1-, 2-, and 4-bit quantization; the lower graphic shows 8-, 16-, and 32-bit quantization. Rows show classification, clustering, retrieval, and STS. Blue, green, and purple correspond to seeds 1, 2, and 3, respectively.}
    \label{fig:rotation_seed}
\end{figure*}

Figure~\ref{fig:rotation_seed} shows
the three score curves for each task family and bit-width. They nearly
coincide for most settings, particularly as the retained dimensionality
increases. Classification shows more visible variation at small dimensions
and low bit-widths. The most aggressive STS setting, two dimensions with
1-bit quantization, also varies across seeds. Thus, seed sensitivity is
mainly confined to strongly compressed settings in this experiment.

Table~\ref{tab:rotation_seed_std} reports the standard deviation across the
three seeds for each task family and bit-width. For each entry, we select the
target dimension with the largest observed score range across seeds. The
table therefore summarizes a variable setting rather than averaging
standard deviations over dimensions. The largest values occur for STS at
1 bit and classification at 1 bit, whereas the standard deviations for
8-, 16-, and 32-bit settings are small across all four task families.

\begin{table}[!t]
    \centering
    \footnotesize
    \setlength{\tabcolsep}{3pt}
    \begin{tabular}{lrrrrrr}
    \toprule
    Task family & 1 bit & 2 bit & 4 bit & 8 bit & 16 bit & 32 bit \\
    \midrule
    Classification & 1.788 & 0.847 & 0.319 & 0.012 & 0.003 & 0.003 \\
    Clustering     & 0.383 & 0.334 & 0.132 & 0.103 & 0.063 & 0.063 \\
    Retrieval      & 0.466 & 0.766 & 0.443 & 0.248 & 0.205 & 0.205 \\
    STS            & 3.504 & 1.388 & 0.490 & 0.040 & 0.006 & 0.006 \\
    \bottomrule
    \end{tabular}
    \caption{Standard deviation of task-family scores across three random rotation seeds for gte-Qwen2 (0--100 scale). For each task family and bit-width, the reported value is measured at the target dimension with the largest score range across seeds.}
    \label{tab:rotation_seed_std}
\end{table}

\end{document}